\documentclass{article}
\usepackage{acl2023,times}
\usepackage{graphicx}
\usepackage{siunitx}
\usepackage{subcaption}
\usepackage{natbib}
\usepackage{caption}
\usepackage{booktabs}
\usepackage{tabularx}
\usepackage{array}
\usepackage{placeins}
\usepackage{amsmath} 
\usepackage{algorithm}
\usepackage{algpseudocode}
\usepackage{newtxmath}
\usepackage{cleveref}
\usepackage{hyperref}
\usepackage{makecell}
\usepackage{multirow}
\usepackage{listings}
\usepackage{color}
\usepackage{xcolor}
\usepackage{widetable}
\usepackage{pgffor}
\usepackage{float}
\usepackage{subcaption}
\usepackage{rotating}
\usepackage{soul}

\crefname{section}{§}{§§}
\Crefname{section}{§}{§§}
\newcommand{\myMethod}{\textsc{RepoAgent }}

\definecolor{lightgreenshade}{HTML}{bce3bd}
\sethlcolor{lightgreenshade}

\title{RepoAgent: An LLM-Powered Open-Source Framework for Repository-level Code Documentation Generation}

\author{
\textbf{Qinyu Luo}$^{1}$\thanks{\hspace{0.25cm}Indicates equal contribution.},\hspace{0.0cm}
\textbf{Yining Ye}$^{1*}$,\hspace{0.0cm}
\textbf{Shihao Liang}$^{1}$,\hspace{0.0cm}
\textbf{Zhong Zhang}$^{1}$\thanks{\hspace{0.25cm}Corresponding Author.},\hspace{0.05cm}
\textbf{Yujia Qin}$^{1}$,\hspace{0.0cm}
\textbf{Yaxi Lu}$^{1}$,\hspace{0.0cm} 
\textbf{Yesai Wu}$^{1}$,\hspace{0.0cm} \\
\textbf{Xin Cong}$^{1}$,\hspace{0.0cm} 
\textbf{Yankai Lin}$^{2}$,\hspace{0.0cm}
\textbf{Yingli Zhang}$^{3}$,\hspace{0.0cm}
\textbf{Xiaoyin Che}$^{3}$,\hspace{0.0cm}
\textbf{Zhiyuan Liu}$^{1\dag}$,\hspace{0.0cm}
\textbf{Maosong Sun}$^{1}$
\\
\\
$^1$Tsinghua University\hspace{0.5cm} 
$^2$Renmin University of China\hspace{0.5cm}
$^3$Siemens AG.
\\
\texttt{\small qinyuluo123@gmail.com, yeyn2001@gmail.com}
}

\begin{document}

\maketitle

\begin{abstract}
Generative models have demonstrated considerable potential in software engineering, particularly in tasks such as code generation and debugging. However, their utilization in the domain of code documentation generation remains underexplored. To this end, we introduce \textsc{RepoAgent}, a large language model powered open-source framework aimed at proactively generating, maintaining, and updating code documentation. Through both qualitative and quantitative evaluations, we have validated the effectiveness of our approach, showing that \myMethod excels in generating high-quality repository-level documentation. The code and results are publicly accessible at \url{https://github.com/OpenBMB/RepoAgent}.
\end{abstract}

\section{Introduction} \label{sec:intro}

\begin{figure}[t!]
    \centering
\includegraphics[width=1\columnwidth]{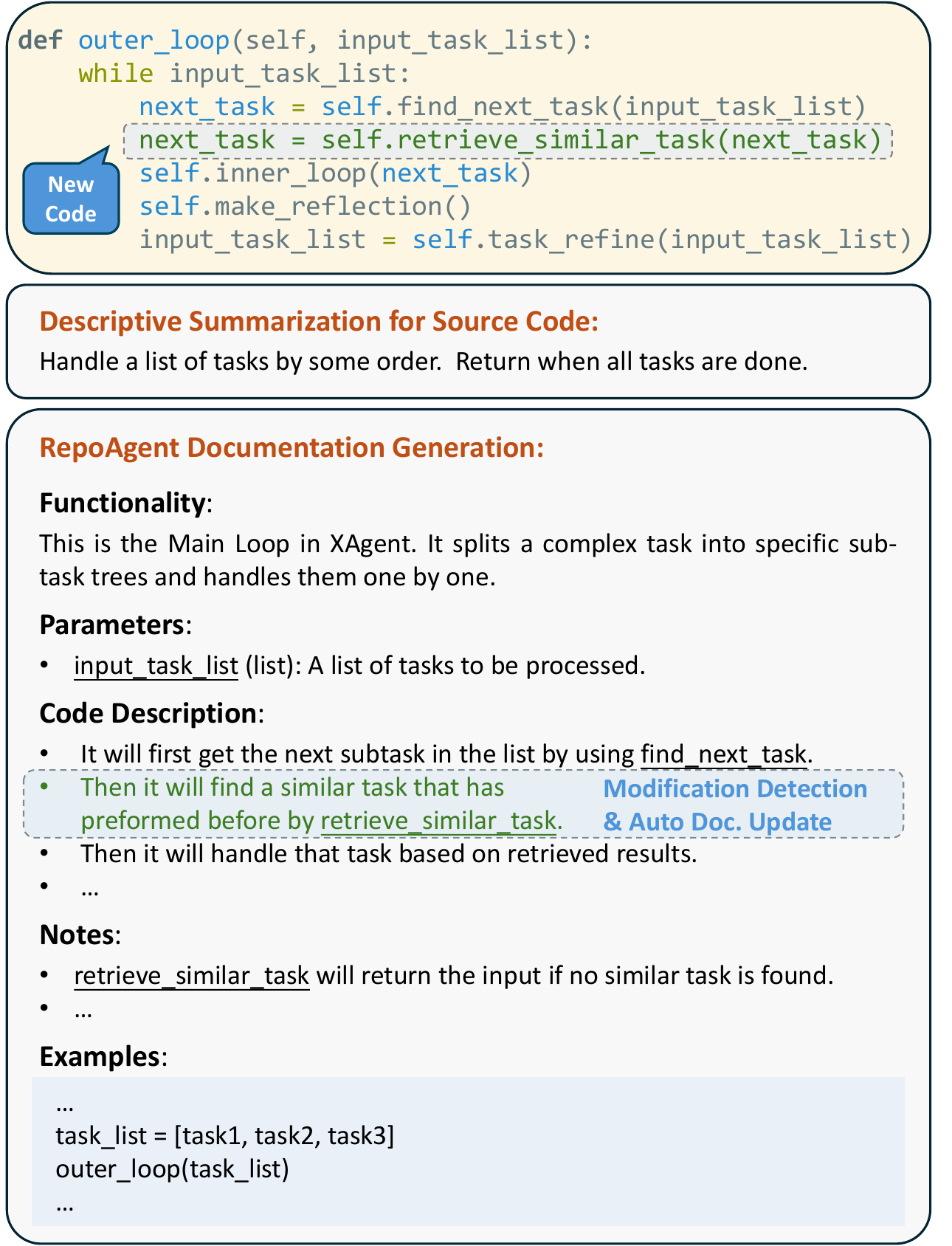}
    \caption{The comparison of code documentation generated by the plain summarization method and the proposed \myMethod.}
    \label{fig:intro_illustrate}
\end{figure}

\begin{figure*}[t!]
    \centering
    \includegraphics[width=\textwidth]{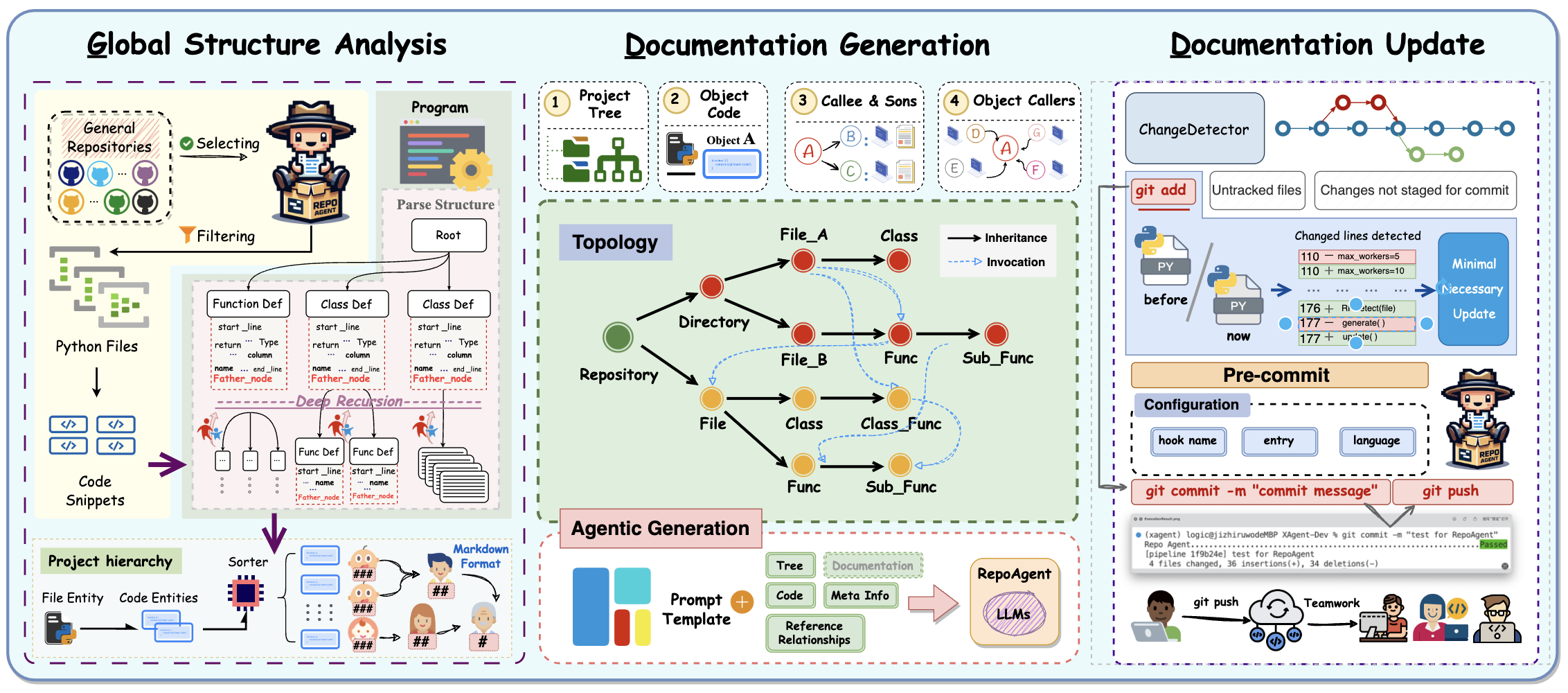}
    \caption{The RepoAgent method consists of Global Structure Analysis, Documentation Generation, and Documentation Update. Each component can be executed independently or packaged as a hook for tooling purposes. When operating as a whole, RepoAgent ensures the capability to construct and maintain documentation for a repository from scratch, elevating documentation to the same level of importance as code, facilitating synchronization and collaboration among teams.}
    \label{fig:method}
\end{figure*}

Developers typically spend approximately 58\% of their time on program comprehension, and high-quality code documentation plays a significant role in reducing this time~\citep{xia2017measuring,Souza2005ASO}. However, maintaining code documentation also consumes a considerable amount of time, money, and human labor~\citep{zhi2015cost}, and not all projects have the resources or enthusiasm to prioritize documentation as their top concern.

To alleviate the burden of maintaining code documentation, early attempts at automatic documentation generation aimed to provide descriptive summaries for source code~\citep{sridhara2010towards,Rai2022ARO,DBLP:conf/kbse/Khan022,DBLP:journals/symmetry/ZhangWZXTGL22}, as illustrated in Figure~\ref{fig:intro_illustrate}. However, they still have significant limitations, particularly in the following aspects: \textbf{(1) Poor summarization}. Previous methods primarily focused on summarizing isolated code snippets, overlooking the dependencies of code within the broader repository-level context. The generated code summaries are overly abstract and fragmented, making it difficult to accurately convey the semantics of the code and compile the code summaries into documentation. \textbf{(2) Inadequate guidance}. Good documentation not only accurately describes the code's functionality, but also meticulously guides developers on the correct usage of the described code~\citep{DBLP:conf/kbse/Khan022, wang2023gdoc}. This includes, but is not limited to, clarifying functional boundaries, highlighting potential misuses, and presenting examples of inputs and outputs. Previous methods still fall short of offering such comprehensive guidance. \textbf{(3) Passive update}. Lehman's first law of software evolution states that a program in use will continuously evolve to meet new user needs~\citep{lehman1980programs}. Consequently, it is crucial for the documentation to be updated in a timely manner to align with code changes, which is the capability that previous methods overlook. Recently, Large Language Models (LLMs) have made significant progress~\citep{openaichatgptblog,openai2023gpt4}, especially in code understanding and generation~\citep{nijkamp2022codegen, li2023starcoder, openaicodex, roziere2023code, xu2023lemur, sun2023prompt, wang2023gdoc, DBLP:conf/kbse/Khan022}. Given these advancements, it is natural to ask: \textbf{Can LLM be used to generate and maintain repository-level code documentation, addressing the aforementioned limitations?}

In this study, we introduce \textsc{RepoAgent}, the first framework powered by LLMs, designed to proactively generate and maintain comprehensive documentation for the entire repository. A running example is demonstrated in \Cref{fig:intro_illustrate}. \myMethod offers the following features: \textbf{(1) Repository-level documentation:} \myMethod leverages the global context to deduce the functional semantics of target code objects within the entire repository, enabling the generation of accurate and semantically coherent structured documentation. \textbf{(2) Practical guidance:} \myMethod not only describes the functionality of the code but also provides practical guidance, including notes for code usage and examples of input and output, thereby facilitating developers' swift comprehension of the code repository. \textbf{(3) Maintenance automation:} \myMethod can seamlessly integrate into team software development workflows managed with Git and proactively take over documentation maintenance, ensuring that the code and documentation remain synchronized. This process is automated and does not require human intervention.

We qualitatively showcased the code documentation generated by \myMethod for real Python repositories. The results reveal that \myMethod is adept at producing documentation of a quality comparable to that created by humans. Quantitatively, in two blind preference tests, the documentation generated by \myMethod was favored over human-authored documentation, achieving preference rates of 70\% and 91.33\% on the Transformers and LlamaIndex repositories, respectively. These evaluation results indicate the practicality of the proposed \myMethod in automatic code documentation generation.

\section{RepoAgent}

\myMethod consists of three key stages: \textbf{global structure analysis}, \textbf{documentation generation}, and \textbf{documentation update}. Figure~\ref{fig:method} shows the overall design of \textsc{RepoAgent}. The \textbf{global structure analysis} stage involves parsing necessary meta information and global contextual relationships from the source code, laying the foundation for \myMethod to infer the functional semantics of the target code. In the \textbf{documentation generation} stage, we have designed a sophisticated strategy that leverages the parsed meta information and global contextual relationships to prompt the LLM to generate fine-grained documentation that is of practical guidance. In the \textbf{documentation update} stage, \myMethod utilizes Git tools to track code changes and update the documentation accordingly, ensuring that the code and documentation remain synchronized throughout the entire project lifecycle.

\subsection{Global Structure Analysis}
\label{subsec:Global Structure Analysis}
An essential prerequisite for generating accurate and fine-grained code documentation is a comprehensive understanding of the code structure. To achieve this goal, we proposed a project tree, a data structure that maintains all code objects in the repository while preserving their semantic hierarchical relationships. Firstly, we filter out all non-Python files within the repository. For each Python file, we apply Abstract Syntax Tree (AST) analysis~\citep{zhang2019novel} to recursively parse the meta information of all Classes and Functions within the file, including their type, name, code snippets, etc. These Classes and Functions associated with their meta information are used as the atomic objects for documentation generation. It is worth noting that the file structures of most well-engineered repositories have reflected the functional semantics of code. Therefore, we first utilize it to initialize the project tree, whose root node represents the entire repository, middle nodes and leaf nodes represent directories and Python files, respectively. Then, we add the parsed Classes and Functions as new leaf nodes (or sub-trees) to the corresponding Python file nodes to form the final project tree.

Beyond the code structure, the reference relationships within the code, as a form of important global contextual information, can also assist the LLM in identifying the functional semantics of the code. Also, references to a target function can be considered natural in-context learning examples~\citep{wei2023chainofthought} to teach the LLM to use the target function, thereby helping generate documentation that is of practical guidance. We consider two types of reference relationships: \texttt{Caller} and \texttt{Callee}. We use the Jedi library\footnote{\url{https://github.com/davidhalter/jedi} Extensible to programming languages other than Python by replacing code parsing tools.} to extract all bi-directional reference relationships in the repository, and then ground them to the corresponding leaf nodes in the project tree. The project tree augmented with the reference relationships forms a Directed Acyclic Graph\footnote{We simply ignored circular dependencies to avoid loops, as most of these situations may have bugs.} (DAG). 
%We perform top-to-bottom topological sorting on this DAG. The sorting result is used to determine the order of documentation generation. This ensures that the child nodes of each node, as well as the nodes it references, have their documentation generated before it.

\subsection{Documentation Generation}
\label{subsec:Documentation Generation}

\begin{figure}[!t]
    \centering
    \includegraphics[width=\columnwidth]{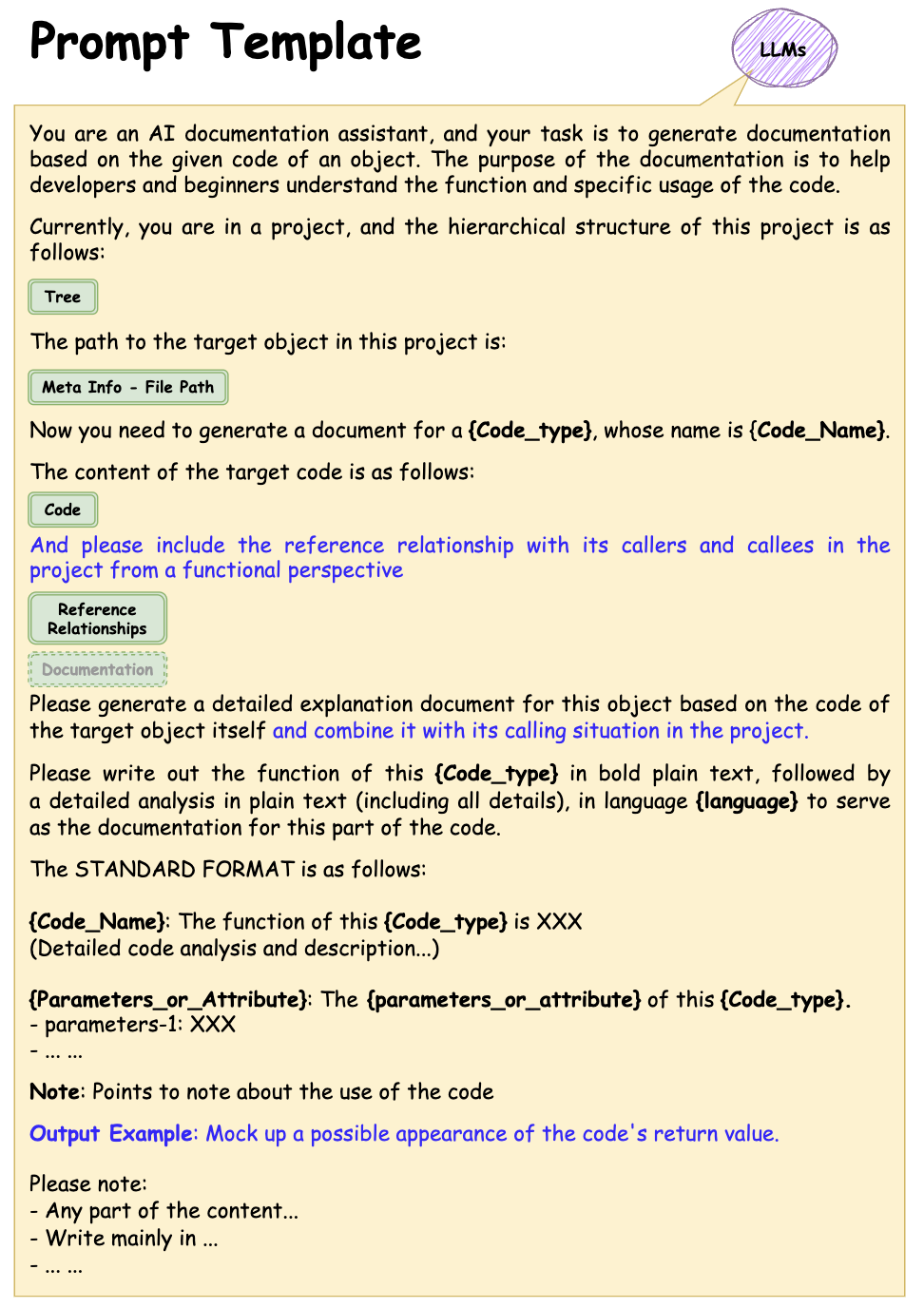}
    \caption{Prompt template used for documentation generation, some details are omitted. Variables within the braces are assigned according to different objects. The blue parts are dynamically filled based on the Meta Info of different objects, enriching the documentation content according to the object characteristics. The Documentation within the dashed boxes can be dynamically utilized according to the program settings. If the documentation information is not used, the program may not execute in topological order.}
    \label{fig:doc_generation_prompt}
\end{figure}

\begin{figure*}[th!]
  \centering
  \includegraphics[width=1\textwidth]{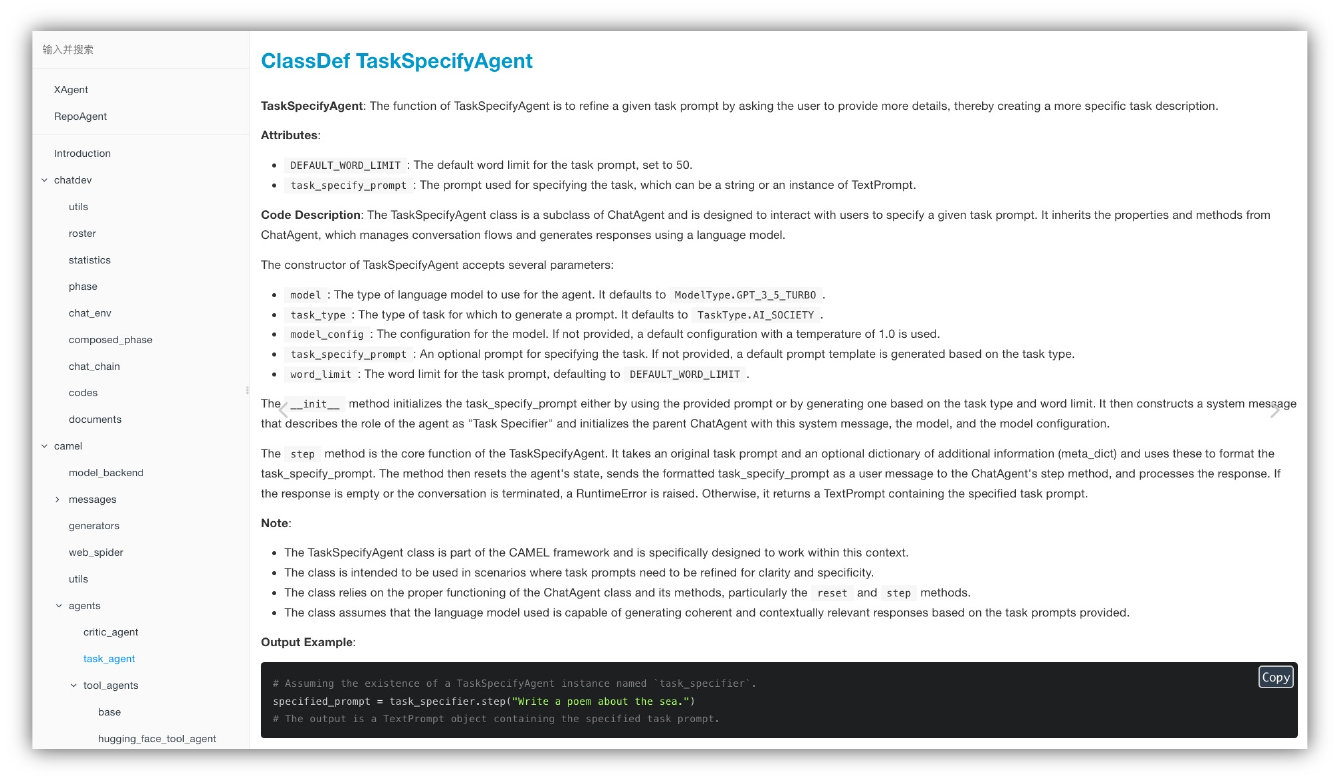} 
  \caption{Demonstration of code documentation generated by \myMethod for the ChatDev repository.}
  \label{fig:Documentation}
\end{figure*}

\myMethod aims to generate fine-grained documentation that is of practical guidance, which includes detailed \textbf{Functionality}, \textbf{Parameters}, \textbf{Code Description}, \textbf{Notes}, and \textbf{Examples}. A backend LLM leverages the parsed meta information and reference relationships from the previous stage to generate documentation with the required structure using a carefully designed prompt template. An illustrative prompt template is shown in Figure~\ref{fig:doc_generation_prompt}, and a complete real-world prompt example is given in~\Cref{lst:prompt_template}. 

The prompt template mainly requires the following parameters: The \textbf{Project Tree} helps \myMethod perceive the repository-level context. The \textbf{Code Snippet} serves as the main source of information for \myMethod to generate the documentation. The \textbf{Reference Relationships} provide semantic invocation relationships between code objects and assist \myMethod in generating guiding notes and examples. The \textbf{Meta Information} indicates the necessary information such as type, name, relative file path of the target object, and is used for post-processing of the documentation. Additionally, we can include the previously generated \textbf{Documentation} of a direct child node of an object as auxiliary information to help code understanding. This is optional, as omitting it can save costs significantly.

\myMethod follows a bottom-to-top topological order to generate documentation for all code objects in the DAG, ensuring that the child nodes of each node, as well as the nodes it references, have their documentation generated before it. After the documentation is generated, \myMethod compiles it into a human-friendly Markdown format. For example, objects of different levels are associated with different Markdown headings (e.g., \#\#, \#\#\#). Finally, \myMethod utilizes GitBook\footnote{\url{https://www.gitbook.com/}} to render the Markdown formatted documentation into a convenient web graphical interface, which enables easy navigation and readability for documentation readers.

\subsection{Documentation Update}
\label{subsec:Documentation Update}

\myMethod supports automatic tracking and updating of documentation through seamless collaboration with Git. The pre-commit hook of Git is utilized to enable \myMethod to detect any code changes and perform documentation updates. After the update, the hook submits both the code and documentation changes, ensuring that the code and documentation remain synchronized. This process is fully automated and does not require human intervention.

Local code changes generally do not affect other code due to the low coupling principle, it is not necessary to regenerate the entire documentation with each minor code update. \myMethod only updates the documentation of affected objects. The updates are triggered when (1) an object's source code is modified; (2) an object's referrers no longer reference it; or (3) an object gets new references. It is worth noting that the update is not triggered when an object's reference objects change, because we adhere to the dependency inversion principle~\citep{martin1996dependency}, which states that high-level modules should not depend on the implementations of low-level modules.

\section{Experiments}

\subsection{Experimental Settings}

We selected 9 Python repositories of varying scales for documentation generation, ranging from less than 1,000 to over 10,000 lines of code. These repositories are renowned for their classic status or high popularity on GitHub, and are characterized by their high-quality code and considerable project complexity. The detailed statistics of the repositories are provided in~\Cref{appendix:imple_details}. We adopted the API-based LLMs \texttt{gpt-3.5-turbo}~\citep{openaichatgptblog} and \texttt{gpt-4-0125}~\citep{openai2023gpt4}, along with the open-source LLMs \texttt{Llama-2-7b} and \texttt{Llama-2-70b}~\citep{DBLP:journals/corr/abs-2307-09288} as backend models for \myMethod. 

\subsection{Case Study}

We use the ChatDev repository~\cite{qian2023communicative} and the \texttt{gpt-4-0125} backend for a case study. The generated documentation is illustrated in~\Cref{fig:Documentation}. Documentation generated by \textsc{RepoAgent} is structured into several parts, starting with a clear, concise sentence that articulates the object's functionality. Following this, the parameters section enumerates all relevant parameters along with their descriptions, aiding developers in understanding how to leverage the provided code. Moreover, the code description section comprehensively elaborates on all aspects of the code, implicitly or explicitly demonstrating the object's role and its associations with other code within the global context. In addition, the notes section further enriches these descriptions by covering usage considerations for the object at hand. Notably, it highlights any logical errors or potential optimization within the code, thereby prompting advanced developers to make modifications. Lastly, if the current object yields a return value, the model will generate an examples section, filled with simulated content to clearly demonstrate the expected output. This is highly advantageous for developers, facilitating efficient code reuse and unit test construction.

\begin{figure}[!t]
    \centering
    \includegraphics[width=1\linewidth]{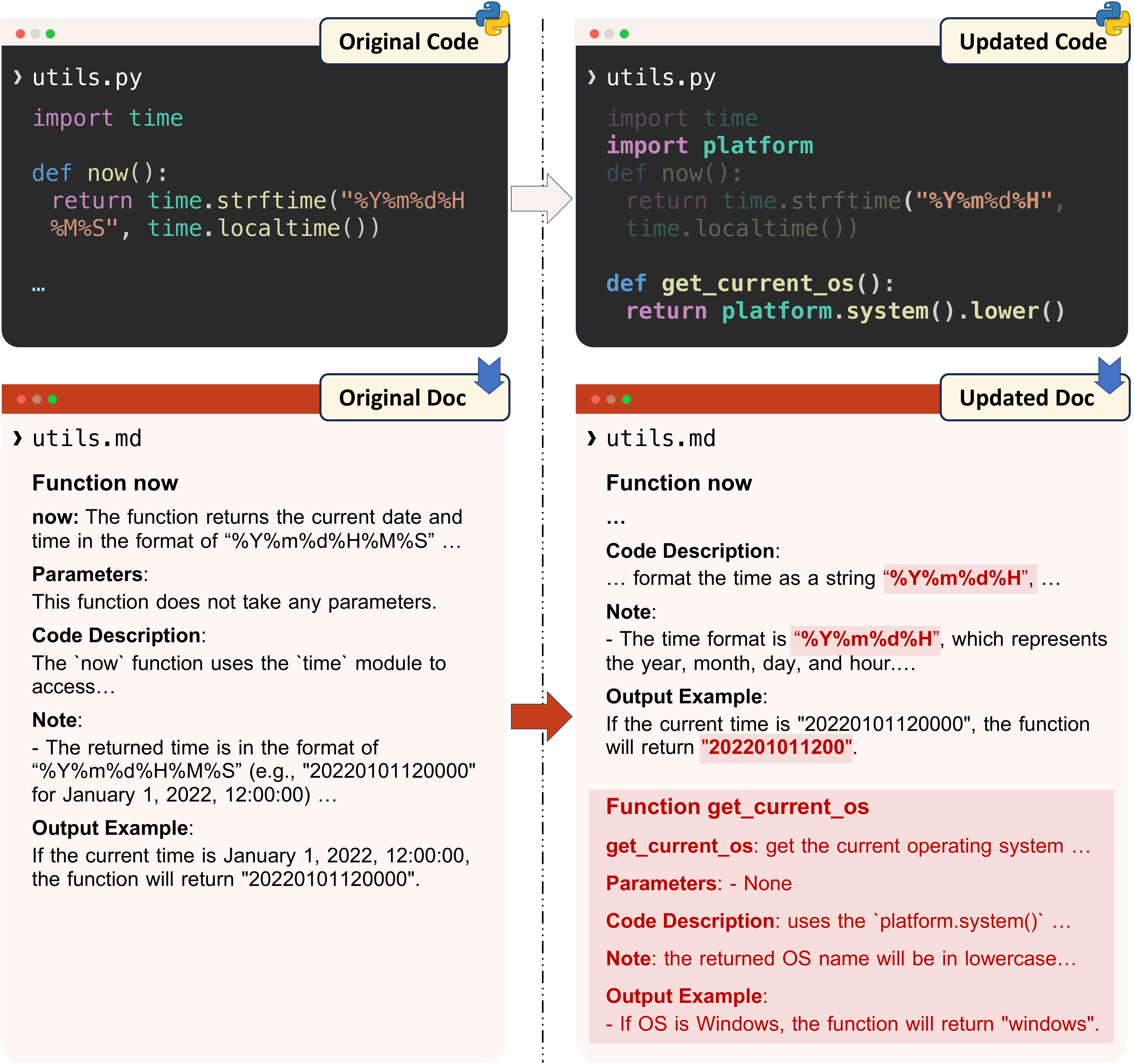}
    \caption{Documentation update for functions of ChatDev.}
    \label{fig:chatdev_case}
\end{figure}

Once the code is changed, the documentation update will be triggered, as illustrated in~\Cref{fig:chatdev_case}. Upon code changes in the staging area, \myMethod identifies affected objects and their bidirectional references, updates documentation for the minimally impacted scope, and integrates these updates into a new Markdown file, which includes additions or global removals of objects' documentation. This automation extends to integrating the pre-commit hook of Git to detect code changes and update documentation, thus seamlessly maintaining documentation alongside project development. Specifically, when code updates are staged and committed, \myMethod is triggered, automatically refreshing the documentation and staging it for the commit. It confirms the process with a ``Passed" indicator, without requiring extra commands or manual intervention, preserving developers' usual workflows.

\subsection{Human Evaluation}
We adopted human evaluation to assess the quality of generated documentation due to the lack of effective evaluation methods. We conducted a preference test to compare human-authored and model-generated code documentation. We randomly sampled 150 pieces of documentation content, including 100 class objects and 50 function-level objects, from both the Transformers and LlamaIndex repositories respectively. Three evaluators were recruited to assess the quality of both documentation sets, with the detailed evaluation protocol outlined in~\Cref{sec:human_eval_protocol}. The results, presented in~\Cref{tab:human_doc and ours doc comparison}, underscore RepoAgent's notable effectiveness in producing documentation that surpasses human-authored content, achieving win rates of $0.70$ and $0.91$, respectively.

\begin{table}[htpb!]
    \centering
    \begin{widetabular}{\columnwidth}{lcccc}
        \toprule
        & Total &  Human & Model & Win Rate \\
        \midrule
        \textbf{Transformers} & 150 & 45 &  105 & \textbf{0.70} \\
        \textbf{LlamaIndex} & 150 & 13  & 137 & \textbf{0.91} \\
        \bottomrule
    \end{widetabular}
    \caption{Results of human preference test on human-authored and model-generated code documentation.}
    \label{tab:human_doc and ours doc comparison}
\end{table}

\subsection{Quantitative Analysis}

\paragraph{Reference Recall.}

We evaluated the models' perception of global context by calculating the recall for identifying reference relationships of code objects. We sampled 20 objects from each of 9 repositories and compared 3 documentation generation methods for their recall in global caller and callee identification. The comparison methods included a machine learning based method that uses LSTM for comment generation~\citep{iyer2016summarizing}, long context concatenation leveraging LLMs with up to 128k context lengths to process entire project codes for identifying calling relationships, single-object generation method that only provides code snippets to LLMs.

\begin{figure}[!htpb]
  \centering
  \includegraphics[width=1\columnwidth]{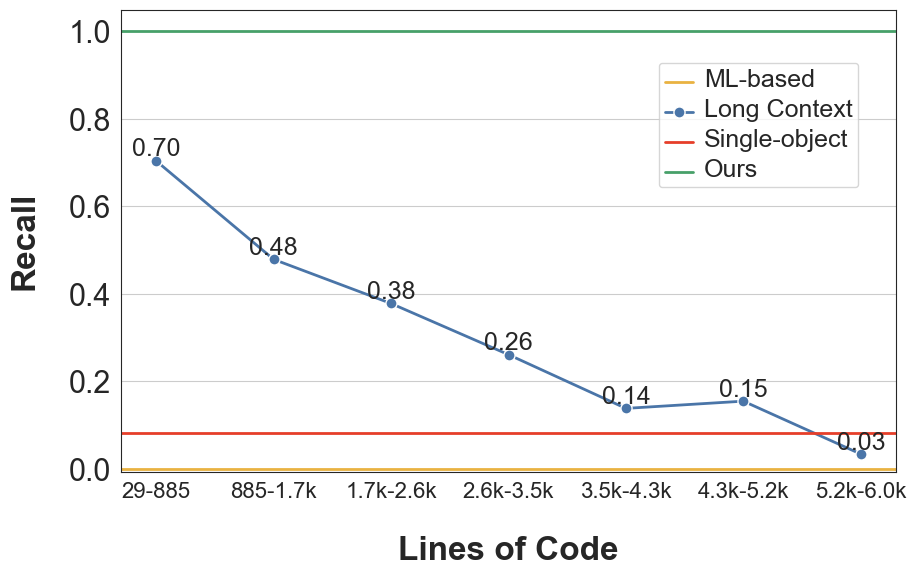} 
  \caption{Recall for identifying reference relationships.}
  \label{fig:4-method Recall}
\end{figure}

Figure \ref{fig:4-method Recall} demonstrates the recall for identifying reference relationships. The machine learning based method is unable to identify reference relationships, whereas the Single-object method partially identifies callees but not callers. The Long Context method, despite offering extensive code content, achieves only partial and non-comprehensive recognition of references, with recall declining as context increases. In contrast, our approach utilizes deterministic tools Jedi and bi-directional parsing to accurately convey global reference relationships, effectively overcoming the scope limitations that other methods encounter in generating repository-level code documentation.

\begin{table*}[!htpb]
    \centering
    \begin{widetabular}{\textwidth}{lcccc}
    \toprule
    \textbf{Repository} &  \texttt{\textbf{Llama-2-7b}}& \texttt{\textbf{Llama-2-70b}} & \texttt{\textbf{gpt-3.5-turbo}} & \texttt{\textbf{gpt-4-0125}} \\
\midrule
     unoconv   &   0.0000& 0.5000 & \textbf{1.0000}&  \textbf{1.0000}   \\
     simdjson  &  0.4298& 0.6336 & \textbf{1.0000}&  0.9644   \\
     greenlet  &  0.5000& 0.7482 & 0.9252&  \textbf{0.9615}   \\
     code2flow & 0.5145& 0.6171 & 0.9735&  \textbf{0.9803}   \\
     AutoGen   &   0.3049& 0.5157 & 0.8633&  \textbf{0.9545}   \\
     AutoGPT   &   0.4243& 0.5611 & 0.8918&  \textbf{0.9527}   \\
     ChatDev   &   0.5387& 0.6980 & 0.9164&  \textbf{0.9695}   \\
     MemGPT    &    0.4582& 0.5729 & 0.9285&  \textbf{0.9911}   \\
     MetaGPT   &   0.3920& 0.5819 & 0.9066&  \textbf{0.9708}   \\
    \bottomrule
    \end{widetabular}
    \caption{Accuracy of identifying function parameters with different LLMs as backends.}
    \label{tab:param_attribute accuracy}
\end{table*}

\begin{figure}[!htpb]
    \centering
    \includegraphics[width=0.945\linewidth]{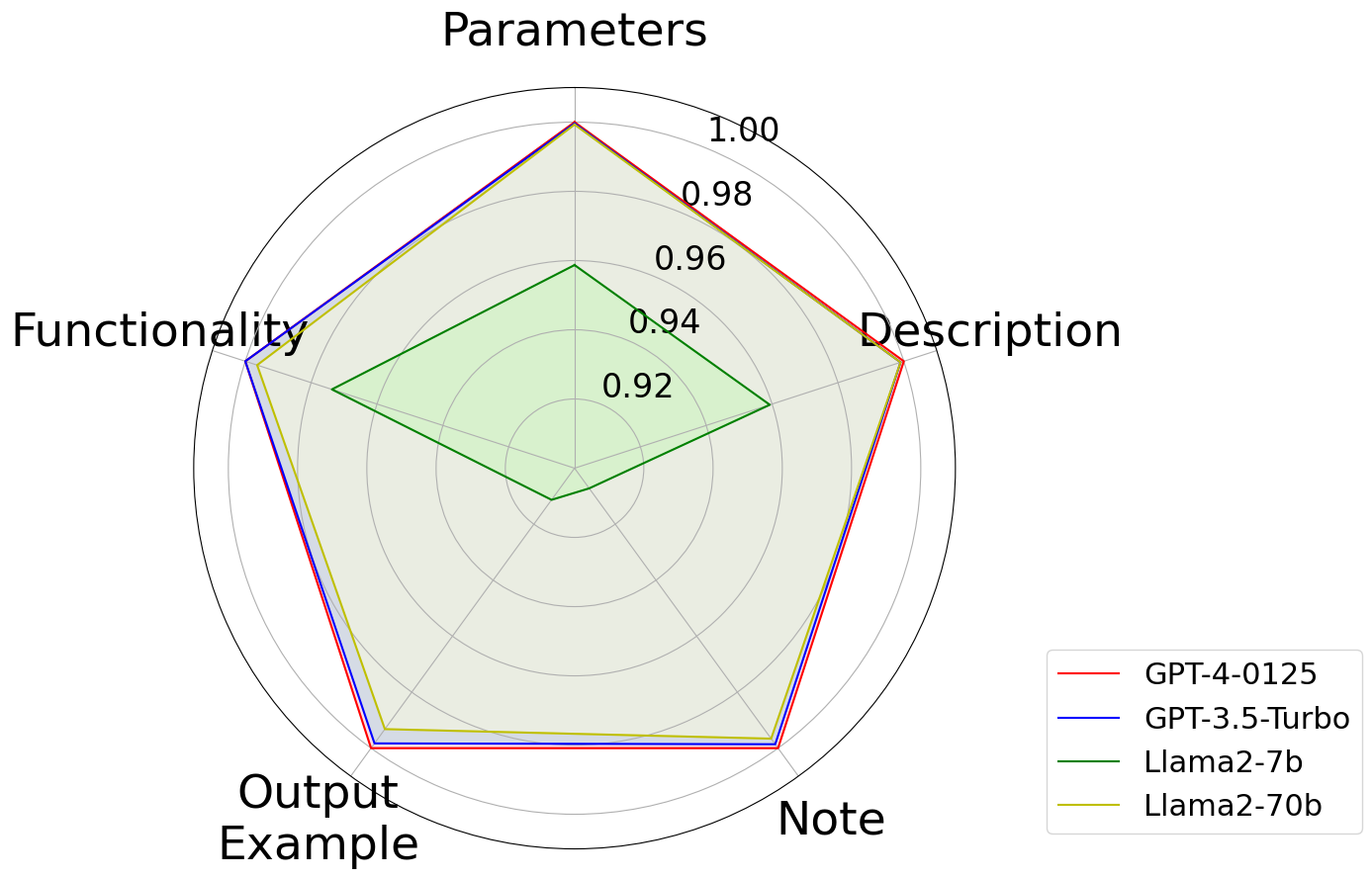}
    \caption{Format alignment accuracy of different LLMs.}
    \label{fig:radar}
\end{figure}

\paragraph{Format Alignment.}
Adherence to the format is critical in documentation generation. The generated documentation should consist of 5 basic parts, where the \textit{Examples} is dynamic, depending on whether the code object has a return value or not. We evaluated the ability of LLMs to adhere to the format using all 9 repositories, the results are shown in~\Cref{fig:radar}. Large models like GPT series and \texttt{Llama-2-70b} perform very well in format alignment, while the small model \texttt{Llama-2-7b} performs poorly, especially in terms of the examples.

\paragraph{Parameter Identification.}

We further evaluated the models' capability to identify parameters on all 9 repositories, the results are shown in~\Cref{tab:param_attribute accuracy}. It is worth noting that we report the accuracy instead of recall, because models may hallucinate non-existent parameters, which should be taken into account. As seen in the table, the GPT series significantly outperforms the LLaMA series in parameter identification, and \texttt{gpt-4-0125} performs the best.

\section{Related Work}
\label{sec:related_work}

\paragraph{Code Summarization.} The field focuses on generating succinct, human-readable code summaries. Early methods were rule-based or template-driven~\cite{haiduc2010use, sridhara2010towards, moreno2013automatic, rodeghero2014improving}. With advancements in machine learning, learning-based approaches like CODE-NN, which utilize LSTM units, emerged for summary creation~\cite{iyer2016summarizing}. The field further evolved with attention mechanisms and transformer models, significantly enhancing the ability to model long-range dependencies~\cite{allamanis2016convolutional, vaswani2017attention}, indicating a shift towards more context-aware and flexible summarization techniques.

\paragraph{LLM Development.} The development and application of LLMs have revolutionized both NLP and software engineering fields. Initially, the field was transformed by masked language models like BERT~\cite{devlin2018bert}, followed by advancements in encoder-decoder models, such as the T5 series~\cite{raffel2020exploring}, and auto-regressive models like the GPT series~\cite{radford2018improving}. Auto-regressive models, notable for their sequence generation capabilities, have been effectively applied in code generation~\cite{nijkamp2022codegen, li2023starcoder, openaicodex, roziere2023code, xu2023lemur}, code summarization~\cite{sun2023prompt}, and documentation generation~\cite{wang2023gdoc, DBLP:conf/kbse/Khan022}, highlighting their versatility in programming and documentation tasks. Concurrently, LLM-based agents have become ubiquitous~\cite{xagent2023,qin2024toolllm,GitAgent,ProAgent,Tool-learning}, especially in software engineering ~\cite{chen2023agentverse, qian2023communicative, hong2023metagpt}, facilitating development through role-play and the automatic generation of agents~\cite{wu2023autogen}, thereby enhancing repository-level code understanding, generation and even debugging~\cite{tian2024debugbench}. With the development of LLM-based agents, repository-level documentation generation become solvable as an agent task.

\section{Conclusion and Discussion}

In this paper, we introduce \textsc{RepoAgent}, an open source framework designed to generate fine-grained repository-level code documentation, facilitating improved team collaboration. The experimental results suggest that \myMethod is capable of generating and proactively maintaining high-quality documentation for the entire project. \myMethod is expected to free developers from this tedious task, thereby improving their productivity and innovation potential.

In future work, we consider how to effectively utilize this tool and explore ways to apply \myMethod to a broader range of downstream applications in the future. To this end, we believe that chatting can serve as a natural tool for establishing a communication bridge between code and humans. Currently, by employing our approach with retrieval-augmented generation, which combines code, documentation, and reference relationships, we have achieved preliminary results in what we called ``Chat~With~Repo'', which marks the advent of a novel coding paradigm.

\section*{Limitations}

\paragraph{Programming Language Limitations.}

\myMethod currently relies on the Jedi reference recognition tool, limiting its applicability exclusively to Python projects. A more versatile, open-source tool that can adapt to multiple programming languages would enable broader adoption across various codebases, which will be addressed in future iterations.

\paragraph{Requirement for Human Oversight.}

AI-generated documentation may still require human review and modification to ensure its accuracy and completeness. Technical intricacies, project-specific conventions, and domain-specific terminology may necessitate manual intervention to enhance the quality of generated documentation.

\paragraph{Dependency on Language Model Capabilities.}

The performance of \myMethod significantly depends on the backend LLMs and associated technologies. Although current results have shown promising progress with API-based LLMs like GPT series, the long-term stability and sustainability of using open-source models still require further validation and research.

\paragraph{Lack of Standards for Evaluation.}

It is difficult to establish a unified quantitative evaluation method for the professionalism, accuracy, and standardization of generated documentation.  Furthermore, it is worth noting that the academic community currently lacks benchmarks and datasets of exemplary human documentation. Additionally, the subjective nature of documentation further limits current methods in terms of quality assessment.

\section*{Broader Impact}

\paragraph{Enhancing Productivity and Innovation.}
\myMethod automates the generation, update and maintenance of code documentation, which is traditionally a time-consuming task for developers. By freeing developers from this burden, our tool not only enhances productivity but also allows more time for creative and innovative work in software development.

\paragraph{Improving Software Quality and Collaboration.}
High-quality documentation is crucial for understanding, using, and contributing to software projects, facilitating developers' swift comprehension of projects. \myMethod's ability ensures long-term high consistency in code documentation. We posit that integrating \myMethod closely with the project development process can introduce a new paradigm for standardizing and making repositories more readable. This, in turn, is expected to stimulate active community contributions and rapid development  with higher overall quality of software projects.

\paragraph{Educational Benefits.}
\myMethod can serve as an educational tool by providing clear and consistent documentation for codebases, making it easier for students and novice programmers to learn software development practices and understand complex codebases.

\paragraph{Bias and Inaccuracy.}
While \myMethod aims to generate high-quality documentation, there's a potential risk of generating biased or inaccurate content due to model hallucination.

\paragraph{Security and Privacy Concerns.}
Currently, \myMethod mainly relies on remote API-based LLMs, which will have the opportunity to access users' code data. This may raise security and privacy concerns, especially for proprietary software. Ensuring data protection and secure handling of the code is crucial.

\section*{Acknowledgments}

We appreciate the suggestions and assistance from all the fellow students and friends in the community, including Arno (Bangsheng Feng), Guo Zhang, Qiang Guo, Yang Li, Yang Jiao, and others.

\bibliography{main}
\bibliographystyle{acl_natbib}

\appendix

\onecolumn
\section{Appendix: Experimental Details}

\subsection{Implementation Details}
\label{appendix:imple_details}

Table \ref{tab:repo_stat} presents the detailed statistics of the selected repositories and the token costs associated with the production of initial documentation. These repositories are sourced from both well-established, highly-starred projects and newly emerged, top-performing projects on GitHub. Repositories are characterized by their numbers of lines of code, classes and functions. Including global information like the project's directory structure and bidirectional references results in very long prompts (as detailed in Appendix \ref{sec:complete prompts}). Despite this, the resulting documentation is thorough yet concise, typically ranging between 0.4k and 1k tokens in length.

During the actual generation process, we addressed the issue of varying text lengths across different models. When using models with shorter context lengths (e.g., \texttt{gpt-3.5-turbo} and the LLaMA series), \myMethod adaptively switches to models with larger context lengths (e.g., \texttt{gpt-3.5-16k} or \texttt{gpt-4-32k}) based on the current prompt's length, to cope with the token overhead of incorporating global perspectives. In cases where even these models' limits are exceeded, \myMethod truncates the content by simplifying the project's directory structure and removing bidirectional reference code before reinitiating the documentation generation task. Such measures are infrequent when employing models with the longest contexts (128k), such as \texttt{gpt-4-1106} or \texttt{gpt-4-0125}. This dynamic scheduling strategy, combined with variable network conditions, may influence token consumption. Nevertheless, \myMethod ensures the integrity of the documentation while striving for cost-effectiveness to the greatest extent.
 
\subsection{Settings}

\subsubsection{Technical Environment}

All experiments were conducted within a Python 3.11.4 environment. The system had CUDA 11.7 installed and was equipped with 8 {NVIDIA A100 40GB} GPUs.  

\subsubsection{Human Evaluation Protocol}
\label{sec:human_eval_protocol}

We recruited three human evaluators to assess the code documentation generated by \textsc{RepoAgent}, and instructed all human evaluators to give an overall evaluation considering a set of evaluation criteria shown in \Cref{tab:criteria}. We randomly sampled 150 pieces of documentation from the repository. Subsequently, each human evaluator was assigned 50 pairs of documentation, each containing one human-authored and one model-generated documentation. The human evaluators were required to select the better documentation for each pair.

\begin{table*}[ht]
  \centering
  \begin{widetabular}{\textwidth}{lcccccc}
    \toprule
    \thead{\textbf{Repository}} & \thead{\textbf{Model}} & \thead{\textbf{Prompt Tokens}} & \thead{\textbf{Completion}\\ \textbf{Tokens}} & \thead{\textbf{Class}\\ \textbf{Numbers}} & \thead{\textbf{Function}\\ \textbf{Numbers}} & \thead{\textbf{Code Lines}} \\
    
    \midrule
    \multirow{4}{*}{unoconv} & \texttt{gpt-4-0125} & \multirow{2}{*}{4020} & 2550 & \multirow{4}{*}{0} & \multirow{4}{*}{1} & \multirow{4}{*}{$\leq$1k} \\
    & \texttt{gpt-3.5-turbo} &  & 2743 & & & \\
    & \texttt{Llama-2-7b} & \multirow{2}{*}{1180} & 2916 & & & \\
    & \texttt{Llama-2-70b} &  & 437 & & & \\

    \midrule
    \multirow{4}{*}{simdjson} & \texttt{gpt-4-0125} & \multirow{2}{*}{45344} & 35068 & \multirow{4}{*}{6} & \multirow{4}{*}{55} & \multirow{4}{*}{$\leq$ 1k} \\
    & \texttt{gpt-3.5-turbo} &  & 29736 & & & \\
    & \texttt{Llama-2-7b} & \multirow{2}{*}{49615} & 27562 & & & \\
    & \texttt{Llama-2-70b} &  & 32961 & & & \\

    \midrule
    \multirow{4}{*}{greenlet} & \texttt{gpt-4-0125} & \multirow{2}{*}{86587} & 79113 & \multirow{4}{*}{59} & \multirow{4}{*}{319} & \multirow{4}{*}{1k $\leq$ 10k} \\
    & \texttt{gpt-3.5-turbo} &  & 260464 & & & \\
    & \texttt{Llama-2-7b} & \multirow{2}{*}{33177} & 31561 & & & \\
    & \texttt{Llama-2-70b} &  & 225595 & & & \\

    \midrule
    \multirow{4}{*}{code2flow} & \texttt{gpt-4-0125} & \multirow{2}{*}{185511} & 134462 & \multirow{4}{*}{51} & \multirow{4}{*}{257} & \multirow{4}{*}{1k $\leq$ 10k} \\
    & \texttt{gpt-3.5-turbo} &  & 234101 & & & \\
    & \texttt{Llama-2-7b} & \multirow{2}{*}{354574} & 431761 & & & \\
    & \texttt{Llama-2-70b} &  & 187835 & & & \\

    \midrule
    \multirow{4}{*}{AutoGen} & \texttt{gpt-4-0125} & \multirow{2}{*}{4939388} & 516975 & \multirow{4}{*}{64} & \multirow{4}{*}{590} & \multirow{4}{*}{1k $\leq$ 10k} \\
    & \texttt{gpt-3.5-turbo} &  & 288609 & & & \\
    & \texttt{Llama-2-7b} & \multirow{2}{*}{889050} & 630139 & & & \\
    & \texttt{Llama-2-70b} &  & 410256 & & & \\

    \midrule
    \multirow{4}{*}{AutoGPT} & \texttt{gpt-4-0125} & \multirow{2}{*}{4116296} & 888223 & \multirow{4}{*}{318} & \multirow{4}{*}{1170} & \multirow{4}{*}{$\geq$ 10k} \\
    & \texttt{gpt-3.5-turbo} &  & 799380 & & & \\
    & \texttt{Llama-2-7b} & \multirow{2}{*}{1838425} & 1893041 & & & \\
    & \texttt{Llama-2-70b} &  & 927946 & & & \\

    \midrule
    \multirow{4}{*}{ChatDev} & \texttt{gpt-4-0125} & \multirow{2}{*}{2021168} & 602474 & \multirow{4}{*}{183} & \multirow{4}{*}{729} & \multirow{4}{*}{$\geq$ 10k} \\
    & \texttt{gpt-3.5-turbo} &  & 519226 & & & \\
    & \texttt{Llama-2-7b} & \multirow{2}{*}{1122400} & 946131 & & & \\
    & \texttt{Llama-2-70b} &  & 531838 & & & \\

    \midrule
    \multirow{4}{*}{MemGPT} & \texttt{gpt-4-0125} & \multirow{2}{*}{628482} & 345109 & \multirow{4}{*}{74} & \multirow{4}{*}{478} & \multirow{4}{*}{$\geq$ 10k} \\
    & \texttt{gpt-3.5-turbo} &  & 234101 & & & \\
    & \texttt{Llama-2-7b} & \multirow{2}{*}{742591} & 740783 & & & \\
    & \texttt{Llama-2-70b} &  & 352940 & & & \\

    \midrule
    \multirow{4}{*}{MetaGPT} & \texttt{gpt-4-0125} & \multirow{2}{*}{154364} & 111159 & \multirow{4}{*}{291} & \multirow{4}{*}{885} & \multirow{4}{*}{$\geq$ 10k} \\
    & \texttt{gpt-3.5-turbo} &  & 134101 & & & \\
    & \texttt{Llama-2-7b} & \multirow{2}{*}{1904244} & 2265991 & & & \\
    & \texttt{Llama-2-70b} & & 1009996 & & & \\

    \bottomrule
  \end{widetabular}
  \caption{Statistics for the selected repositories and the token consumption for documentation generation. Note that token count calculation varies with each model's tokenizer, rendering direct comparisons between different models impractical.}
  \label{tab:repo_stat}

\end{table*}

\subsubsection{Reference Recall}

The experiment aims to evaluate the model's ability to perceive global context, which is reflected by the recall for identifying reference relationships. The comparison methods are: 

\begin{enumerate}
  \item \textbf{ML-based method.} ~\citet{iyer2016summarizing} utilized traditional machine learning and deep learning methods for generating comments describing the functionality of code objects.
  \item \textbf{Long context concatenation.} The method directly concatenates the code snippets until the context length reaches 128k to let the model discover reference relationships.
  \item \textbf{Single-object generation.} \citet{sun2023prompt} used the GPT-3.5 series to generate documentation by directly feeding code snippets of the target object. We modified the prompt on this basis, adding requirements for outputting the callers and callees.  
\end{enumerate}

Notably, among these methods, only the ML-based approach failed to explicitly or implicitly manifest call relationships in the final document. While it is inherently challenging for a code snippet to discern its invocation throughout the entire repository, the code typically elucidates the current object's calls explicitly. To measure the recall of callers and callees, we enhanced the original documentation by adding information about the calling functions (callers) and the called functions (callees). Then we compared the enriched documentation with our bidirectional reference data from MetaInfo.

For long context concatenation, we randomly selected 20 objects from each of the 9 repositories, culminating in a total of 180 objects. Given the intricate nature of defining context construction criteria for repository-level documentation generation tasks, we circumvented direct concatenation of adjacent and file-adjacent context content. Instead, we formulated negative samples by extracting all objects with reference relationships to fulfill the context length. Leveraging the content of objects and negative sample content, we devised context lengths for the 180 objects, spanning from 29 to 6.0k Code Lines. This approach aimed to optimize the distribution of context lengths while maximizing the utilization of the model's context length. In the case of single-object generation method, we utilized the same pool of 180 objects, providing the model with object source code snippets to generate documentation and elucidate reference relationships.

During the evaluation of both the Long Context Concatenation and Single-object Generation methods, we provided the model with tree-structured hierarchical position information for target objects and their related counterparts. This additional information was intended to help the model in better identifying callers and delineating them in a path form. Despite this assistance, the model's misinterpretations exacerbated as the context length increased, and the Single-object Generation method yielded a substantial amount of speculative information, resulting in unstable and inaccurate caller relationship recognition.

\subsubsection{Format Alignment} 

The experiment evaluates whether the model-generated documentation follows the defined format. LLMs generally excel in instruction following, but the complexity of our task requires models to grasp core intents within lengthy prompts, posing a challenge. We use a one-shot approach with strict output examples, enabling evaluation of model answers through format matching algorithms. Specifically, we mandate that section titles be enclosed in bold symbols, ensure clear divisions between sections, and require contents within sections to be extractable and meaningful.

We observed the shortcomings of open-source models (LLaMA series) in their ability to adhere to formatting. In contrast, the GPT-4 series models excellently achieve format integrity and stability. We also observed behavioral differences between \texttt{gpt-4-0125} and \texttt{gpt-4-1106} models, the former appeared to produce more redundant information.

Format alignment can also be achieved with perfect accuracy using hierarchical or modular generation methods. However, this approach introduces a significant token overhead since each independent module must encompass complete global information and invocation relationships. Current method has demonstrated satisfactory performance on format alignment, meeting human readability standards effectively.

\subsubsection{Parameter Identification}
Accurately identifying and describing parameters or attributes (depending on whether the current object is a function or a class) in code is crucial as it helps readers quickly understand the design logic and usage. We extracted recognized parameters from the Parameters section using a matching pattern: parameters follow a uniform and fixed format, with the parameter name enclosed in code identifiers followed by the parameter's descriptive text.

We organized the extracted parameters into arrays and calculated accuracy by comparing them with the values in the params field (also an array) of the Repository's MetaInfo. It is important to note that we were calculating accuracy here, not recall. This is because some models may hallucinate many nonexistent parameters based on the code snippets. These errors must be taken into consideration, otherwise they will result in biased evaluations.

\begin{table*}[htbp!]
\centering
\renewcommand{\arraystretch}{1.5} 
\begin{tabular}{>{\centering\bfseries}m{0.2\textwidth} m{0.748\textwidth}}

\toprule
\multicolumn{1}{c}{\textbf{Criteria}} & \multicolumn{1}{c}{\textbf{Details}} \\
\midrule
\multirow{5}{*}{\centering Accuracy} & \textbf{Correctness}: Verify if the documentation accurately describes the code's functionality, algorithms, and expected behavior under various conditions. \\
\cmidrule{2-2}
& \textbf{Precision}: Assess whether the documentation provides precise and unambiguous information regarding the code's operations, parameters, and expected outcomes. \\
\cmidrule{2-2}
& \textbf{Alignment with Codebase}: Ensure that the documentation aligns closely with the actual implementation of the code, including any updates or changes made to the codebase. \\
\midrule
\multirow{5}{*}{\centering Completeness} & \textbf{Coverage}: Evaluate if the documentation comprehensively covers all significant aspects of the code, including inputs, outputs, error handling, edge cases, and any potential exceptions. \\
\cmidrule{2-2}
& \textbf{In-depth Explanation}: Determine if the documentation delves into detailed explanations of complex functionalities or algorithms, providing insights into the underlying logic. \\
\cmidrule{2-2}
& \textbf{Documentation of External Dependencies}: Check if the documentation adequately addresses any external libraries, modules, or APIs used within the codebase. \\
\midrule
\multirow{5}{*}{\centering Understandability} & \textbf{Clarity}: Assess the clarity and readability of the documentation, ensuring that it is easily understandable by developers of varying expertise levels. \\
\cmidrule{2-2}
& \textbf{Conciseness}: Determine if the documentation conveys information concisely without unnecessary verbosity or technical jargon that might hinder comprehension. \\
\cmidrule{2-2}
& \textbf{Structured Organization}: Evaluate if the documentation is logically organized, with clear headings, sections, and navigation aids for easy reference and comprehension. \\
\midrule
\multirow{5}{*}{\centering Consistency} & \textbf{Formatting Consistency}: Ensure consistency in the formatting, styling, and layout of the documentation across all sections and pages. \\
\cmidrule{2-2}
& \textbf{Terminology Consistency}: Verify that consistent terminology and naming conventions are used throughout the documentation to maintain coherence and clarity. \\
\cmidrule{2-2}
& \textbf{Style Guide Adherence}: Assess if the documentation adheres to any predefined style guides or conventions established by the project or organization. \\
\midrule
\multirow{3.5}{*}{\centering Relevance} & \textbf{Content Relevance}: Determine if the information provided in the documentation is directly relevant to the code's functionality, purpose, and usage scenarios. \\
\cmidrule{2-2}
& \textbf{Avoidance of Redundancy}: Check for redundancy or repetition within the documentation, eliminating any extraneous or irrelevant details that do not contribute to understanding the code. \\
\midrule
\multirow{5.5}{*}{\centering Examples and Usage} & \textbf{Code Samples}: Evaluate if the documentation includes sufficient code samples, snippets, or examples to illustrate the usage and implementation of key functionalities. \\
\cmidrule{2-2}
& \textbf{Use Cases}: Assess if the documentation provides real-world use cases or scenarios where the code can be applied, demonstrating its practical utility and versatility. \\
\cmidrule{2-2}
& \textbf{Step-by-Step Instructions}: Determine if the documentation offers clear, step-by-step instructions or tutorials for integrating, configuring, and utilizing the code in different environments or applications. \\
\bottomrule
\end{tabular}
\caption{Detailed criteria for human evaluation.}
\label{tab:criteria}
\end{table*}

\onecolumn

\definecolor{mygreen}{rgb}{0,0.6,0}
\definecolor{mygray}{rgb}{0.5,0.5,0.5}
\definecolor{mymauve}{rgb}{0.58,0,0.82}

\lstset{
  frame=tb,
  aboveskip=3mm,
  belowskip=3mm,
  showstringspaces=false,
  columns=flexible,
  basicstyle={\small\ttfamily},
  numbers=none,
  breaklines=true,
  breakatwhitespace=true,
  tabsize=3,
  breakindent=0pt,
  keywordstyle=\color{blue},  
  commentstyle=\color{mygreen},  
  stringstyle=\color{mymauve},  % Add this line
  escapeinside={(*@}{@*)},  % Add this line
}

\section{Appendix: More Cases of Generated Documentation}

\subsection{Documentation Showcases}
In this section, we showcase additional generated documentation to validate the practical application of \myMethod. The included images are direct screenshots from the documentation of two open-source projects, ChatDev and AutoGen. Our intent is to provide readers with a detailed and panoramic view of how our method is utilized in real-world scenarios, thereby offering a deeper understanding of its effectiveness and versatility.

% ChatDev
\begin{figure}[htpb!]
\centering
\begin{subfigure}[b]{\textwidth}
    \centering
    \includegraphics[width=\textwidth]{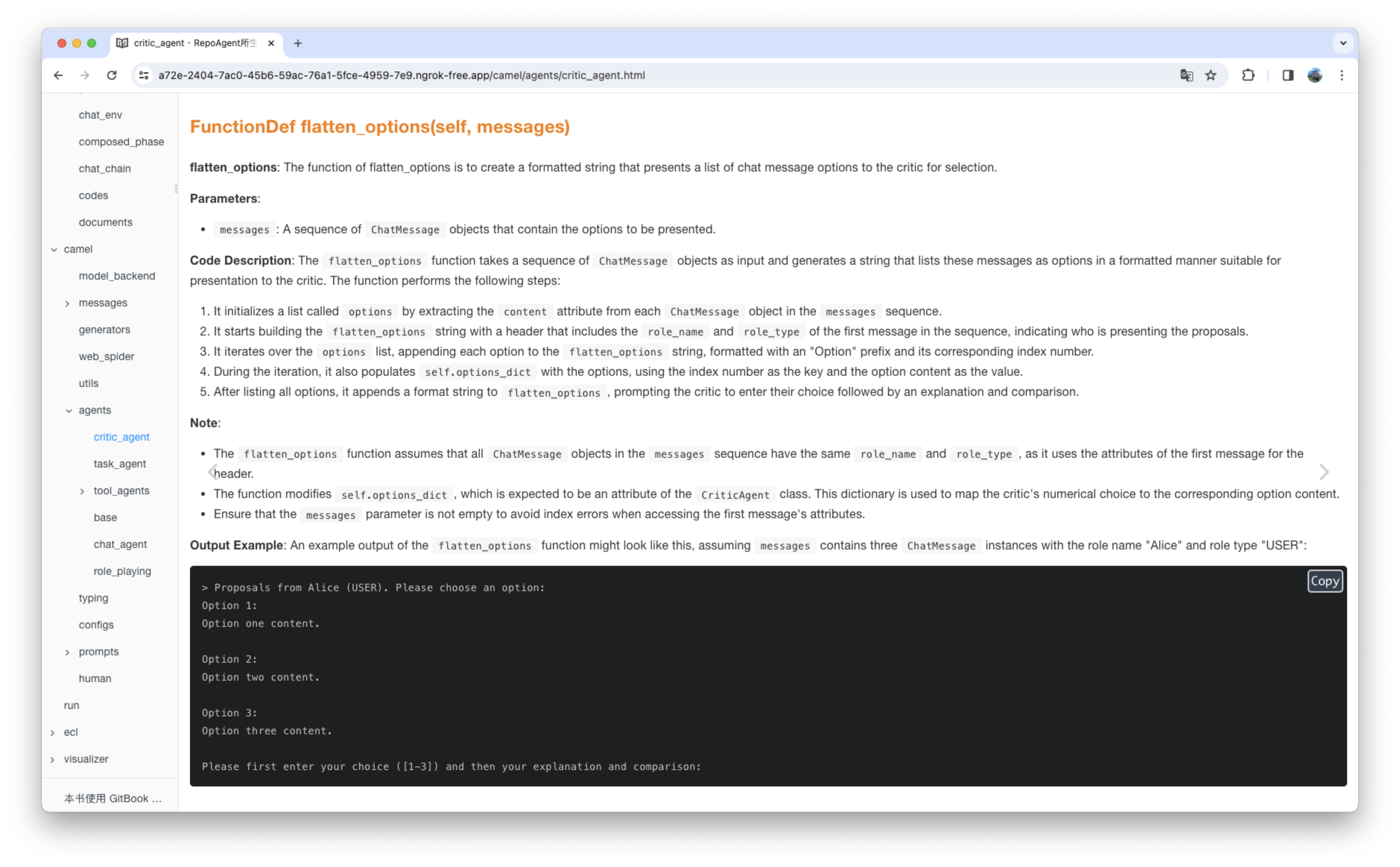}
        \subcaption{}
\end{subfigure}
\\ \vspace{-0mm}
\begin{subfigure}[b]{\textwidth}
    \centering
    \includegraphics[width=\textwidth]{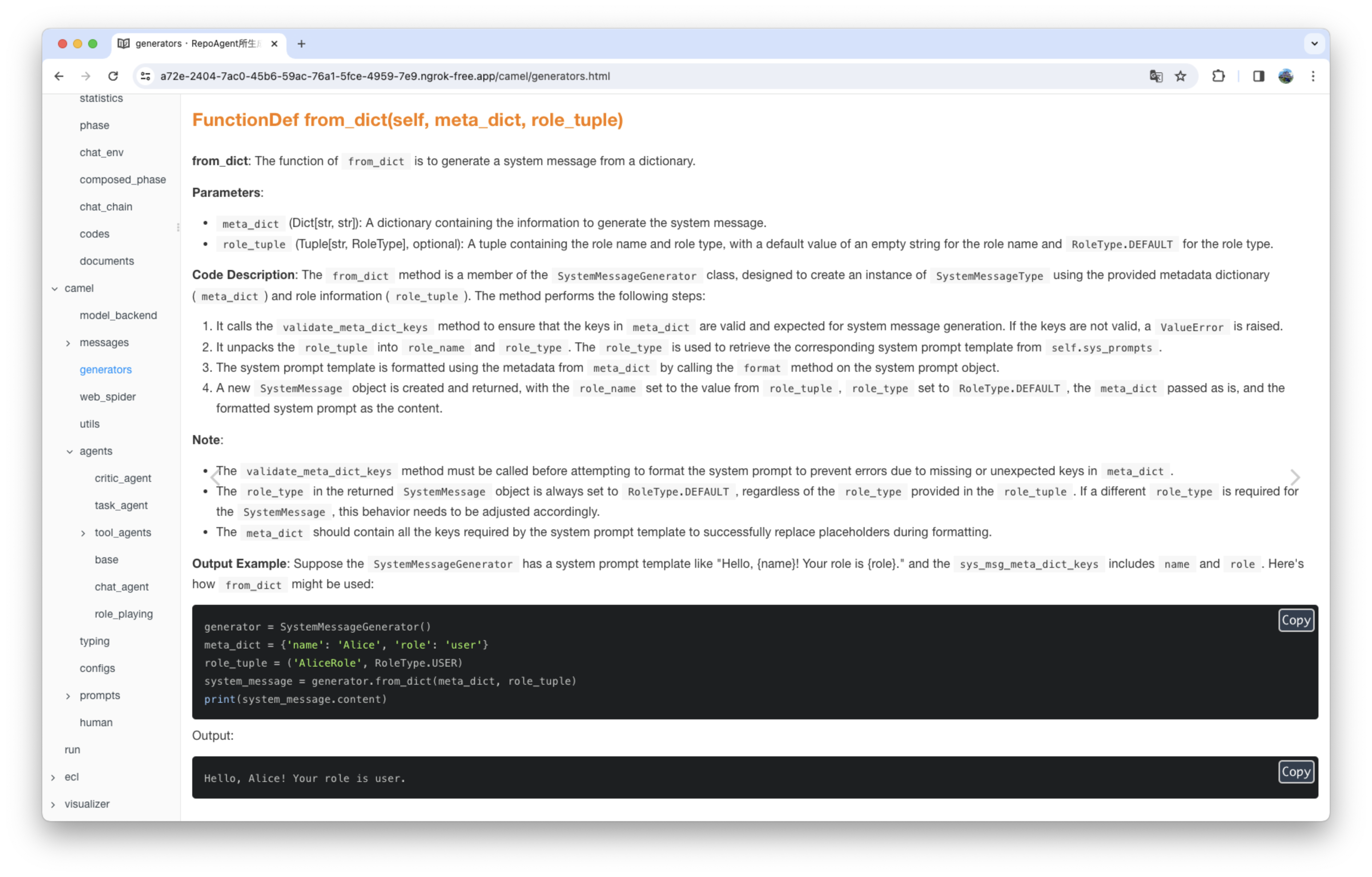}
        \subcaption{}
\end{subfigure}
\end{figure}

\begin{figure}[H] \ContinuedFloat
\centering
\begin{subfigure}[b]{\textwidth}
    \centering
    \includegraphics[width=\textwidth]{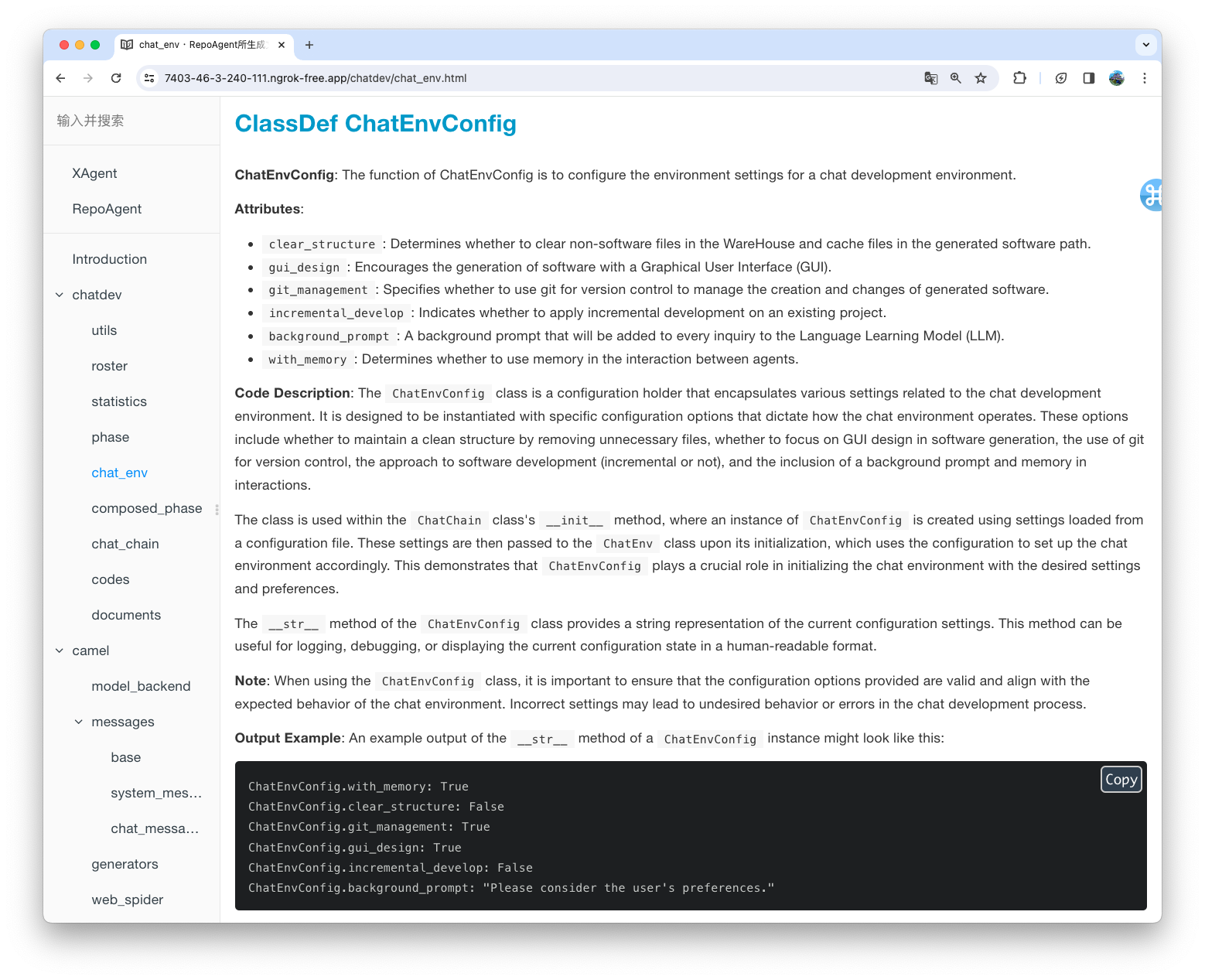}
        \subcaption{}
\end{subfigure}
\\ \vspace{-0mm}
\begin{subfigure}[b]{\textwidth}
    \centering
    \includegraphics[width=\textwidth]{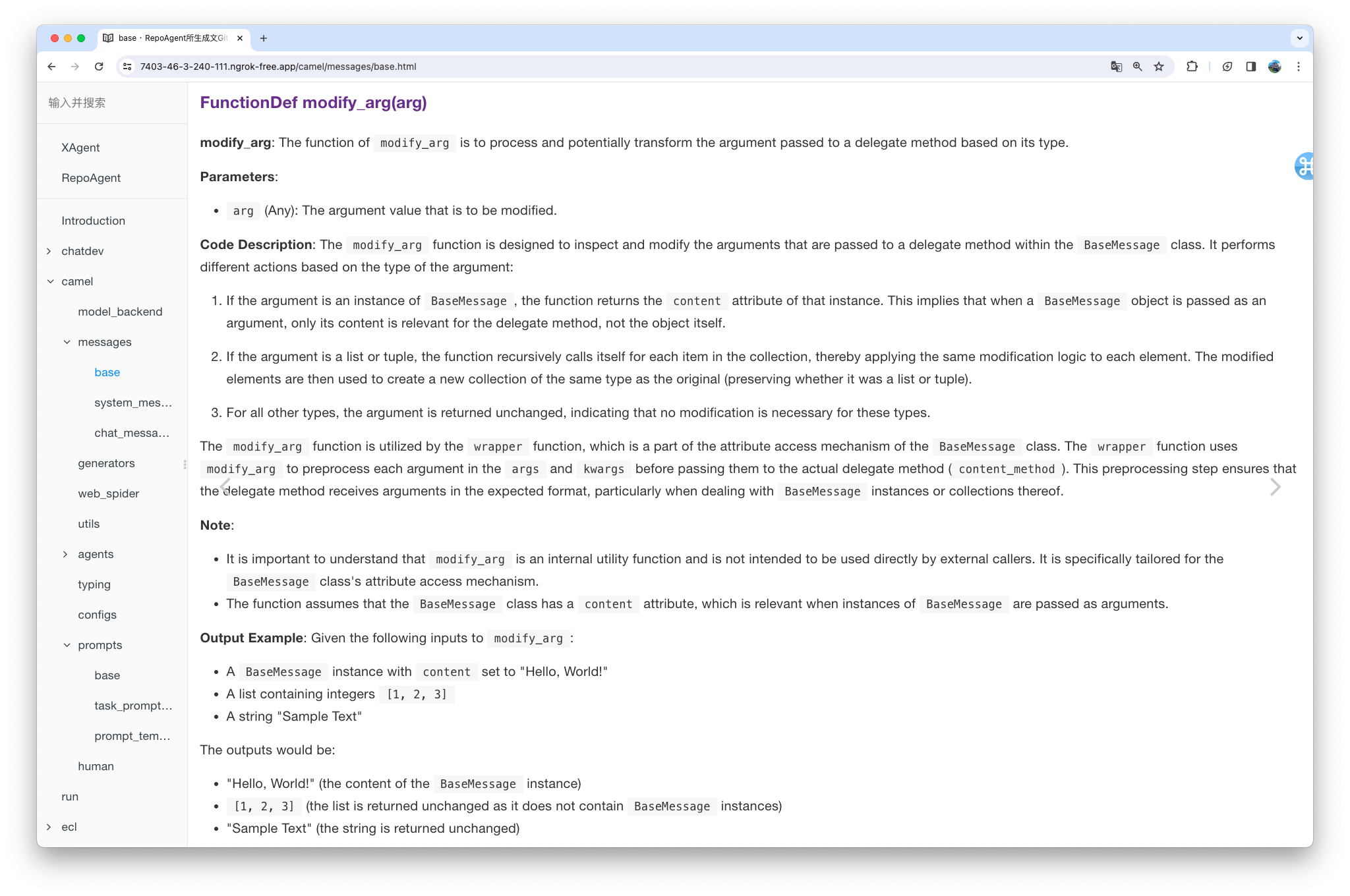}
        \subcaption{}
\end{subfigure}
\end{figure}

\begin{figure}[H] \ContinuedFloat
\centering
\begin{subfigure}[b]{\textwidth}
    \centering
    \includegraphics[width=\textwidth]{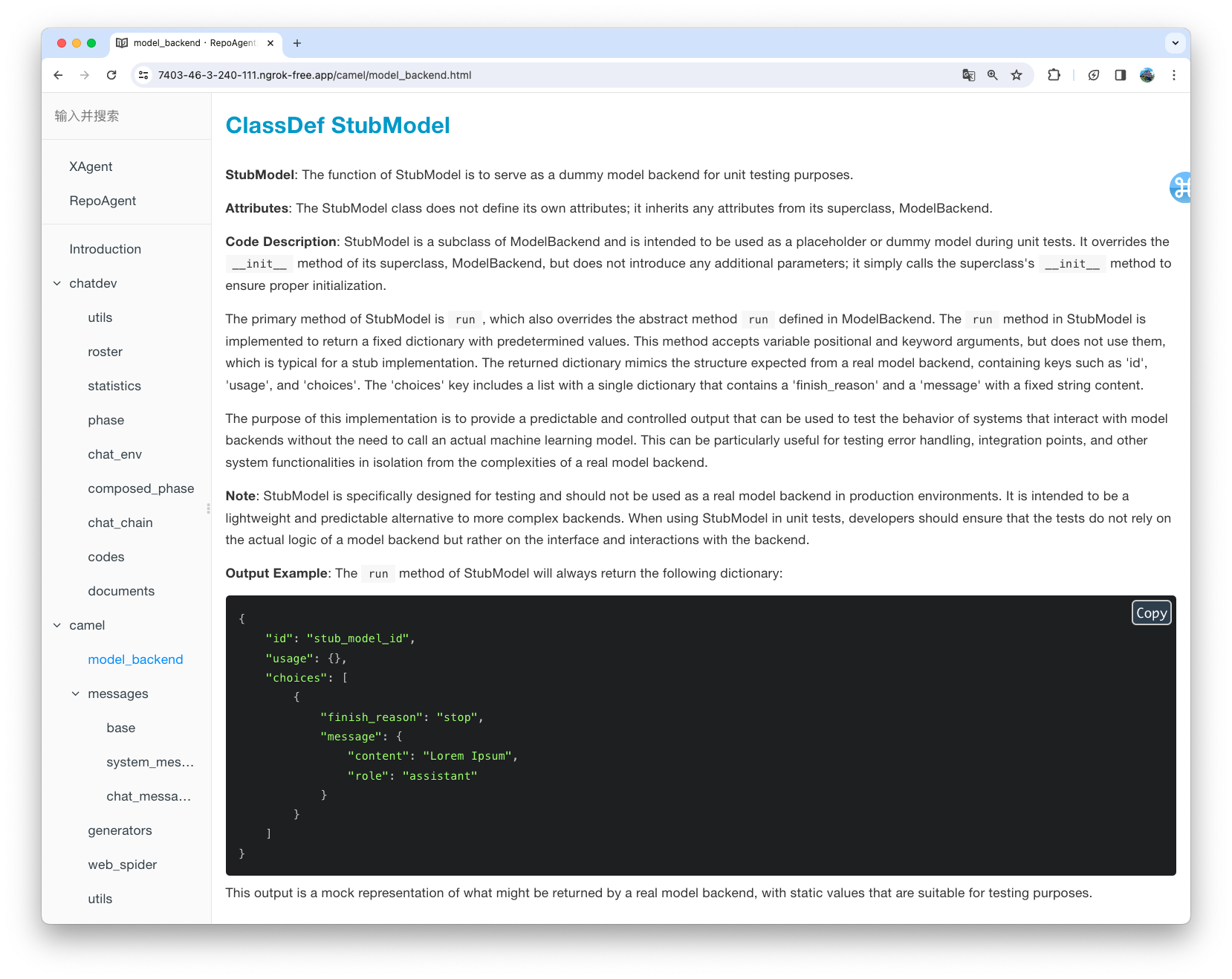}
    \subcaption{}
\end{subfigure}
\\
\begin{subfigure}[b]{\textwidth}
    \centering
    \includegraphics[width=\textwidth]{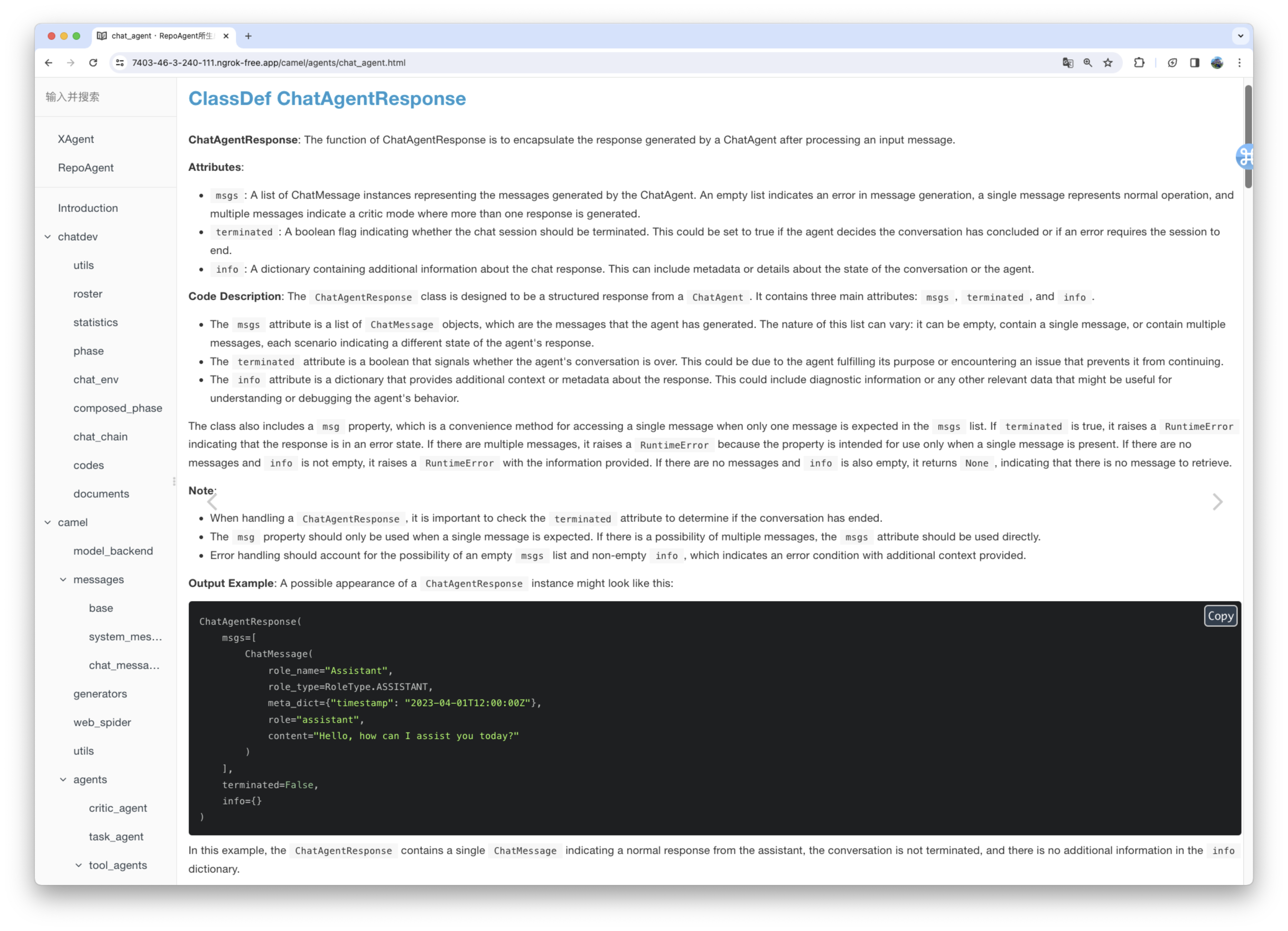}
    \subcaption{}
\end{subfigure}
\end{figure}

\begin{figure}[H] \ContinuedFloat
\centering
\begin{subfigure}[b]{\textwidth}
    \centering
    \includegraphics[width=\textwidth]{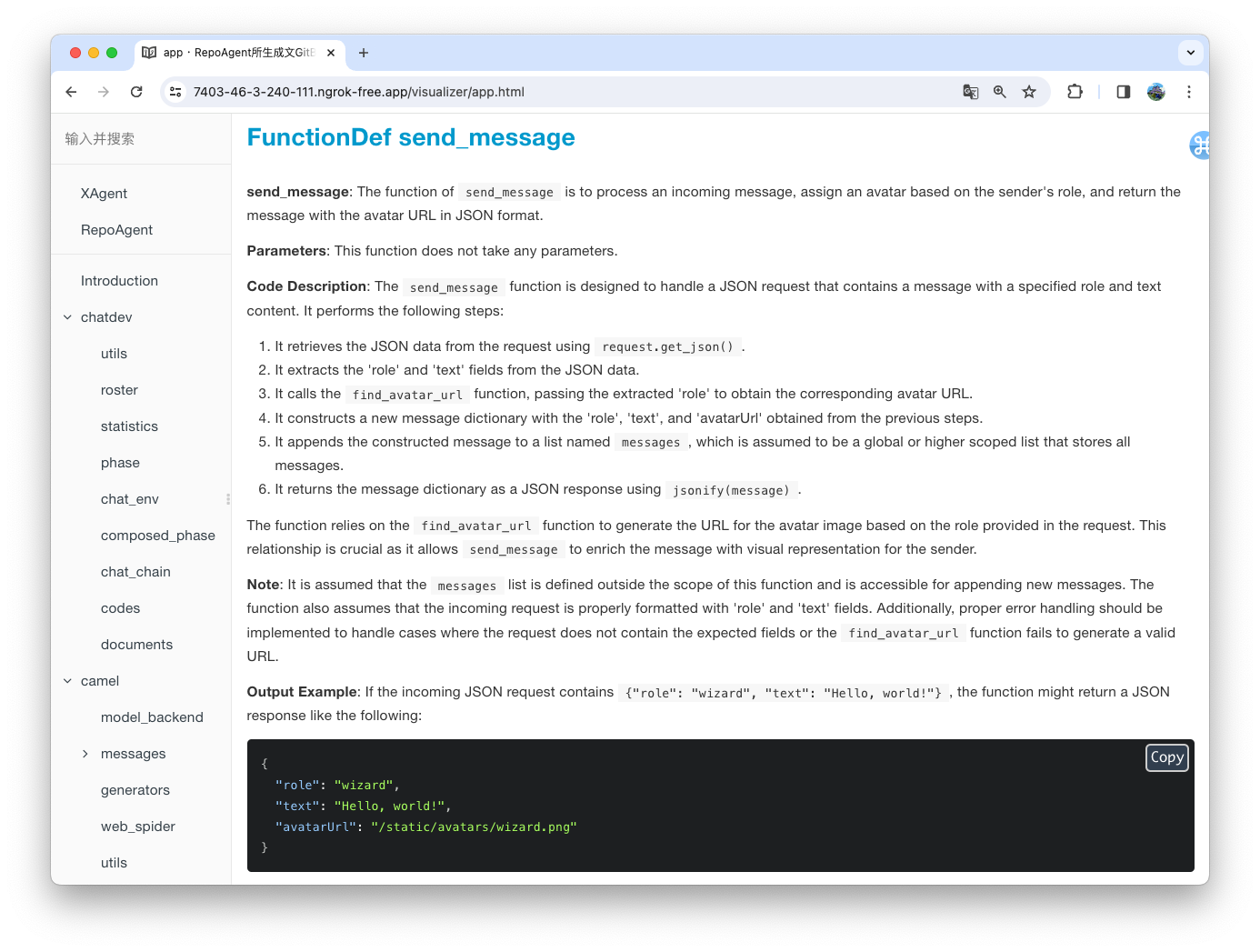
    }
    \subcaption{}
\end{subfigure}
\\
\begin{subfigure}[b]{\textwidth}
    \centering
    \includegraphics[width=\textwidth]{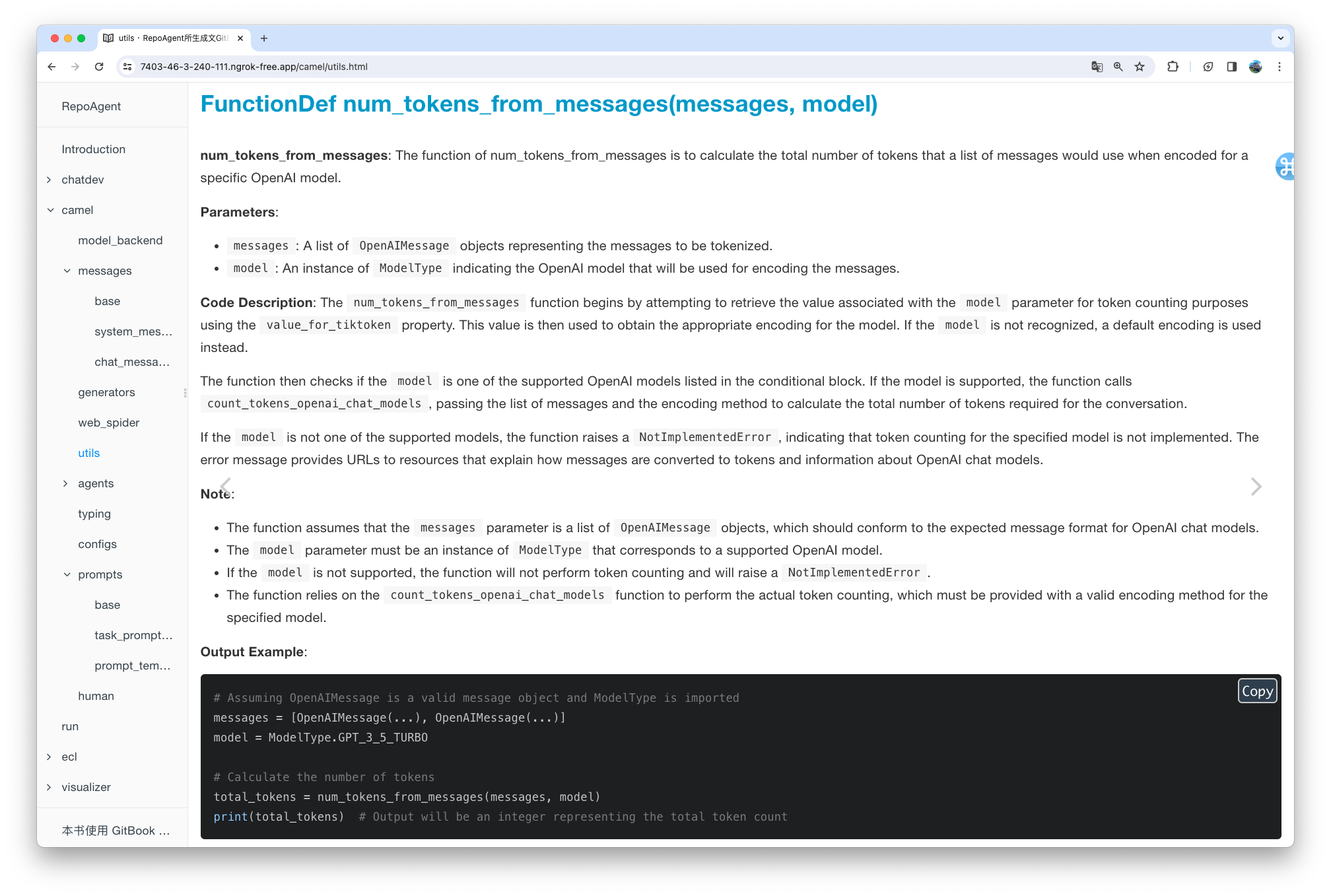}
    \subcaption{}
\end{subfigure}
\caption{Samples of code documentation generated by \myMethod for the ChatDev repository.}
\label{fig:sample_docs_chatdev}
\end{figure}

\begin{center}
    \begin{sideways}%[htbp]
         \begin{minipage}{\paperwidth}
            \includegraphics[width=\paperwidth,keepaspectratio]{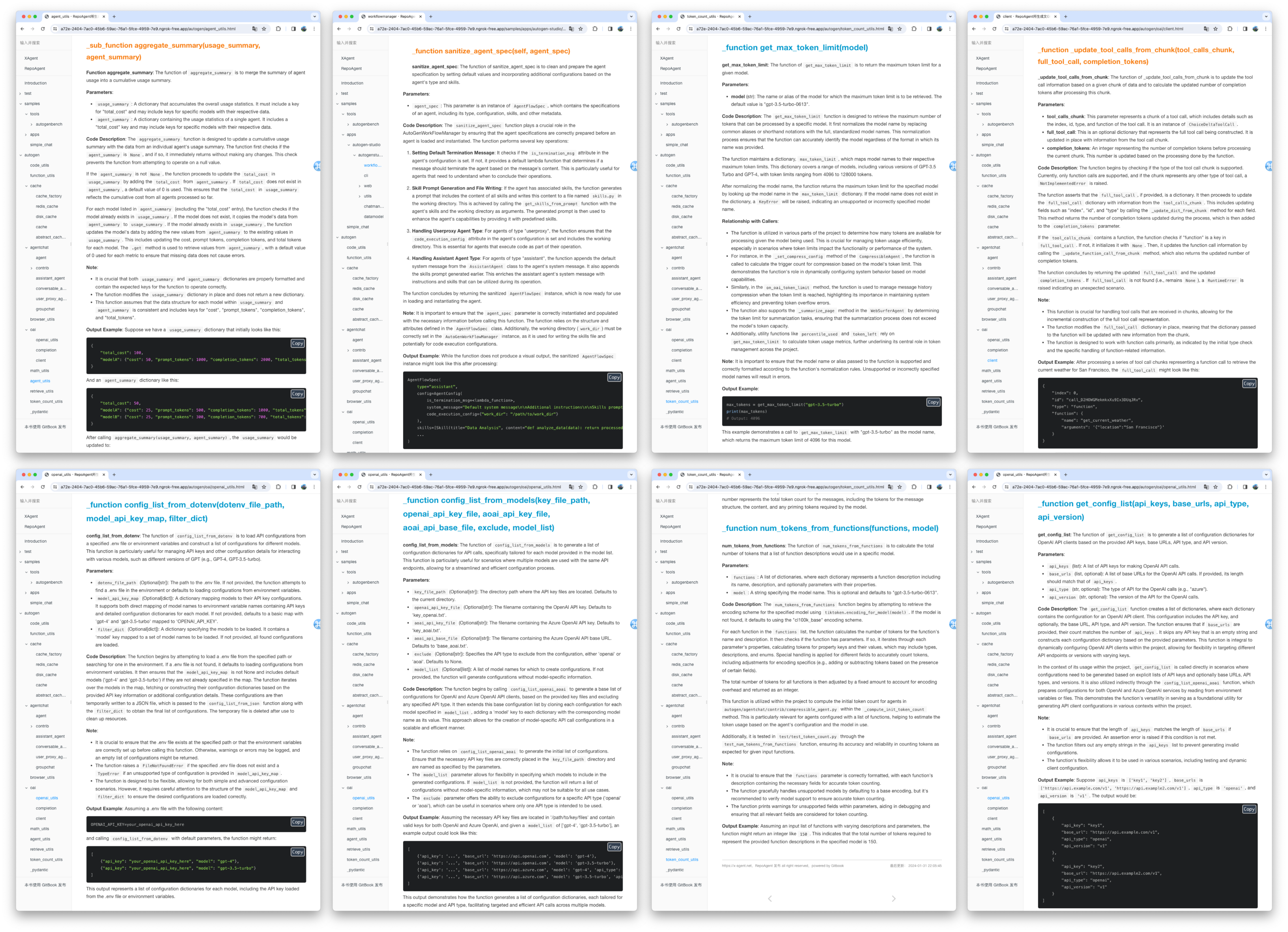}
            \captionof{figure}{Samples of code documentation generated by \myMethod for the AutoGen repository.}
         \label{fig:sample_docs_autogen}
         \end{minipage}
    \end{sideways}
\end{center}

\newpage
\section{Appendix: Full Prompts}
\label{sec:complete prompts}

\subsection{A full prompt of ask\_user method in AutoGPT}

\begin{lstlisting}[label={lst:prompt_template},escapechar=|]
You are an AI documentation assistant, and your task is to generate documentation based on the given code of an object. The purpose of the documentation is to help developers and beginners understand the function and specific usage of the code.

|\hl{Currently, you are in a project, and the related hierarchical structure of this project is as follows (The current object is marked with an *):}|
autogpts
    autogpt
        autogpt
            commands
                user_interaction.py
                    ask_user
                        *ask_user


|\hl{The path of the document you need to generate in this project is:}|
autogpts/autogpt/autogpt/commands/user_interaction.py/ask_user.

Now you need to generate a document for a |\hl{Function}|, whose name is |\hl{"ask\_user"}|.

|\hl{The content of the code is as follows:}|

async def ask_user(question: str, agent: Agent) -> str:
    print(f"\nQ: {question}")
    resp = await clean_input(agent.legacy_config, "A:")
    return f"The user's answer: '{resp}'"


|\hl{As you can see, the code calls the following objects, their code and docs are as following:}|

|\hl{OBJ\_NAME:}| clean_input
|\hl{OBJ\_PATH:}| autogpts/autogpt/autogpt/app/utils.py/clean_input

|\hl{Document:}| 
**Function Name**: clean_input

**Parameters**:
- `config`: An instance of the `Config` class, which holds the configuration settings for the application.
- `prompt`: A string that represents the prompt to be displayed to the user. It defaults to an empty string if not provided.

**Code Description**:
The `clean_input` function is an asynchronous function designed to process user input in a chat-enabled application environment. It performs several key operations based on the application's configuration and the plugins that are enabled.

1. **Plugin Handling**: If chat messages are enabled in the application configuration (`config.chat_messages_enabled`), the function iterates through each plugin configured in the application. For each plugin, it checks if the plugin can handle user input (`can_handle_user_input`). If so, it invokes the plugin's `user_input` method with the user's prompt. Depending on the plugin's response, the function may return specific keys (`config.authorise_key` or `config.exit_key`) or the plugin's response directly.

2. **User Input Collection**: If the application is not handling the input through plugins or if chat messages are not enabled, the function logs a debug message indicating that it is waiting for user input via the keyboard. It then asynchronously waits for the user's input, ensuring that the application's signal handler for interrupt signals (`SIGINT`) is not overridden. This is crucial for allowing the application to handle interrupt signals as configured elsewhere in the application, particularly in `autogpt/main.py`.

3. **Interrupt Handling**: The function is equipped to handle a `KeyboardInterrupt` exception, which occurs when the user attempts to interrupt the application (typically by pressing Ctrl+C). In such cases, it logs a message indicating that the user has interrupted AutoGPT and that the application is quitting, and then exits the application with a status code of 0.

**Note**:
- The function's ability to handle plugins dynamically allows for a flexible and extensible architecture where different plugins can be developed and integrated to process user input in various ways.
- The use of asynchronous I/O (`await session.prompt_async`) for collecting user input makes the function suitable for asynchronous applications, ensuring that the application remains responsive.
- Proper handling of `KeyboardInterrupt` ensures a graceful shutdown of the application when the user decides to interrupt the execution.

**Output Example**:
- If a plugin processes the input and responds with "yes", the function might return the value of `config.authorise_key`.
- If the user inputs "Hello" and no plugins are handling the input, the function returns "Hello".
- If the user interrupts the application, no output is returned as the application exits.

|\hl{[Code begin of clean\_input]}|
```
async def clean_input(config: Config, prompt: str = ""):
    try:
        if config.chat_messages_enabled:
            for plugin in config.plugins:
                if not hasattr(plugin, "can_handle_user_input"):
                    continue
                if not plugin.can_handle_user_input(user_input=prompt):
                    continue
                plugin_response = plugin.user_input(user_input=prompt)
                if not plugin_response:
                    continue
                if plugin_response.lower() in [
                    "yes",
                    "yeah",
                    "y",
                    "ok",
                    "okay",
                    "sure",
                    "alright",
                ]:
                    return config.authorise_key
                elif plugin_response.lower() in [
                    "no",
                    "nope",
                    "n",
                    "negative",
                ]:
                    return config.exit_key
                return plugin_response

        # ask for input, default when just pressing Enter is y
        logger.debug("Asking user via keyboard...")

        # handle_sigint must be set to False, so the signal handler in the
        # autogpt/main.py could be employed properly. This referes to
        # https://github.com/Significant-Gravitas/AutoGPT/pull/4799/files/ 3966cdfd694c2a80c0333823c3bc3da090f85ed3#r1264278776
        answer = await session.prompt_async(ANSI(prompt + " "), handle_sigint=False)
        return answer
    except KeyboardInterrupt:
        logger.info("You interrupted AutoGPT")
        logger.info("Quitting...")
        exit(0)

```==========
|\hl{[Code end of clean\_input]}|

|\hl{Also, the code has been called by the following objects, their code and docs are as following:}|

|\hl{OBJ\_NAME:}| execute_step
|\hl{OBJ\_PATH:}| autogpts/autogpt/autogpt/app/agent_protocol_server.py/ AgentProtocolServer/execute_step
|\hl{Document:}| 
None
|\hl{[Code begin of execute\_step]}|
```
    async def execute_step(self, task_id: str, step_request: StepRequestBody) -> Step:
        """Create a step for the task."""
        logger.debug(f"Creating a step for task with ID: {task_id}...")

        # Restore Agent instance
        task = await self.get_task(task_id)
        agent = configure_agent_with_state(
            state=self.agent_manager.retrieve_state(task_agent_id(task_id)),
            app_config=self.app_config,
            llm_provider=self._get_task_llm_provider(task),
        )

        # According to the Agent Protocol spec, the first execute_step request contains
        #  the same task input as the parent create_task request.
        # To prevent this from interfering with the agent's process, we ignore the input
        #  of this first step request, and just generate the first step proposal.
        is_init_step = not bool(agent.event_history)
        execute_command, execute_command_args, execute_result = None, None, None
        execute_approved = False

        # HACK: only for compatibility with AGBenchmark
        if step_request.input == "y":
            step_request.input = ""

        user_input = step_request.input if not is_init_step else ""

        if (
            not is_init_step
            and agent.event_history.current_episode
            and not agent.event_history.current_episode.result
        ):
            execute_command = agent.event_history.current_episode.action.name
            execute_command_args = agent.event_history.current_episode.action.args
            execute_approved = not user_input

            logger.debug(
                f"Agent proposed command"
                f" {execute_command}({fmt_kwargs(execute_command_args)})."
                f" User input/feedback: {repr(user_input)}"
            )

        # Save step request
        step = await self.db.create_step(
            task_id=task_id,
            input=step_request,
            is_last=execute_command == finish.__name__ and execute_approved,
        )
        agent.llm_provider = self._get_task_llm_provider(task, step.step_id)

        # Execute previously proposed action
        if execute_command:
            assert execute_command_args is not None
            agent.workspace.on_write_file = lambda path: self._on_agent_write_file(
                task=task, step=step, relative_path=path
            )

            if step.is_last and execute_command == finish.__name__:
                assert execute_command_args
                step = await self.db.update_step(
                    task_id=task_id,
                    step_id=step.step_id,
                    output=execute_command_args["reason"],
                )
                logger.info(
                    f"Total LLM cost for task {task_id}: "
                    f"${round(agent.llm_provider.get_incurred_cost(), 2)}"
                )
                return step

            if execute_command == ask_user.__name__:  # HACK
                execute_result = ActionSuccessResult(outputs=user_input)
                agent.event_history.register_result(execute_result)
            elif not execute_command:
                execute_result = None
            elif execute_approved:
                step = await self.db.update_step(
                    task_id=task_id,
                    step_id=step.step_id,
                    status="running",
                )
                # Execute previously proposed action
                execute_result = await agent.execute(
                    command_name=execute_command,
                    command_args=execute_command_args,
                )
            else:
                assert user_input
                execute_result = await agent.execute(
                    command_name="human_feedback",  # HACK
                    command_args={},
                    user_input=user_input,
                )

        # Propose next action
        try:
            next_command, next_command_args, raw_output = await agent.propose_action()
            logger.debug(f"AI output: {raw_output}")
        except Exception as e:
            step = await self.db.update_step(
                task_id=task_id,
                step_id=step.step_id,
                status="completed",
                output=f"An error occurred while proposing the next action: {e}",
            )
            return step

        # Format step output
        output = (
            (
                f"`{execute_command}({fmt_kwargs(execute_command_args)})` returned:"
                + ("\n\n" if "\n" in str(execute_result) else " ")
                + f"{execute_result}\n\n"
            )
            if execute_command_args and execute_command != ask_user.__name__
            else ""
        )
        output += f"{raw_output['thoughts']['speak']}\n\n"
        output += (
            f"Next Command: {next_command}({fmt_kwargs(next_command_args)})"
            if next_command != ask_user.__name__
            else next_command_args["question"]
        )

        additional_output = {
            **(
                {
                    "last_action": {
                        "name": execute_command,
                        "args": execute_command_args,
                        "result": (
                            orjson.loads(execute_result.json())
                            if not isinstance(execute_result, ActionErrorResult)
                            else {
                                "error": str(execute_result.error),
                                "reason": execute_result.reason,
                            }
                        ),
                    },
                }
                if not is_init_step
                else {}
            ),
            **raw_output,
        }

        step = await self.db.update_step(
            task_id=task_id,
            step_id=step.step_id,
            status="completed",
            output=output,
            additional_output=additional_output,
        )

        logger.debug(
            f"Running total LLM cost for task {task_id}: "
            f"${round(agent.llm_provider.get_incurred_cost(), 3)}"
        )
        agent.state.save_to_json_file(agent.file_manager.state_file_path)
        return step

```==========
|\hl{[Code end of execute\_step]}|

Please generate a detailed explanation document for this object based on the code of the target object itself and combine it with its calling situation in the project.

Please write out the function of this Function in bold plain text, followed by a detailed analysis in plain text (including all details), in language English to serve as the documentation for this part of the code.

|\hl{The standard format is as follows:}|

|\hl{**ask\_user**:}| The function of ask_user is XXX
|\hl{**parameters**:}| The parameters of this Function.
- parameter1: XXX
- parameter2: XXX
- ...
|\hl{**Code Description**:}| The description of this Function.
(Detailed and CERTAIN code analysis and description...None)
|\hl{**Note**:}| Points to note about the use of the code
|\hl{**Output Example**:}| Mock up a possible appearance of the code's return value.

Please note:
- Any part of the content you generate SHOULD NOT CONTAIN Markdown hierarchical heading and divider syntax.
- Write mainly in the desired language. If necessary, you can write with some English words in the analysis and description to enhance the document's readability because you do not need to translate the function name or variable name into the target language.

Keep in mind that your audience is document readers, so use a deterministic tone to generate precise content and don't let them know you're provided with code snippet and documents. AVOID ANY SPECULATION and inaccurate descriptions! Now, provide the documentation for the target object in English in a professional way.
\end{lstlisting}

\section{Appendix: Chat With Repo}

Moving beyond documentation generation, we are actively exploring how best to use \myMethod and examining its potential for a broader range of downstream applications in the future. We categorize these applications as:

\begin{itemize}
   \item README.md Generation
   \item Automatic Q\&A for Issues and Source Codes
   \item Unit Test Generation
   \item Automated Development of New Features
   \item Repo Level Debugging
   \item Generation of Public Tutorial Documentation
\end{itemize}

We conceptualize ``\textbf{Chat~With~Repo}'' as a unified gateway for these downstream applications, acting as a connector that links \textsc{RepoAgent} to human users and other AI agents. Our future research will focus on adapting the interface to various downstream applications and customizing it to meet their unique characteristics and implementation requirements.

\begin{figure*}[hpb!]
  \centering
  \includegraphics[width=\textwidth]{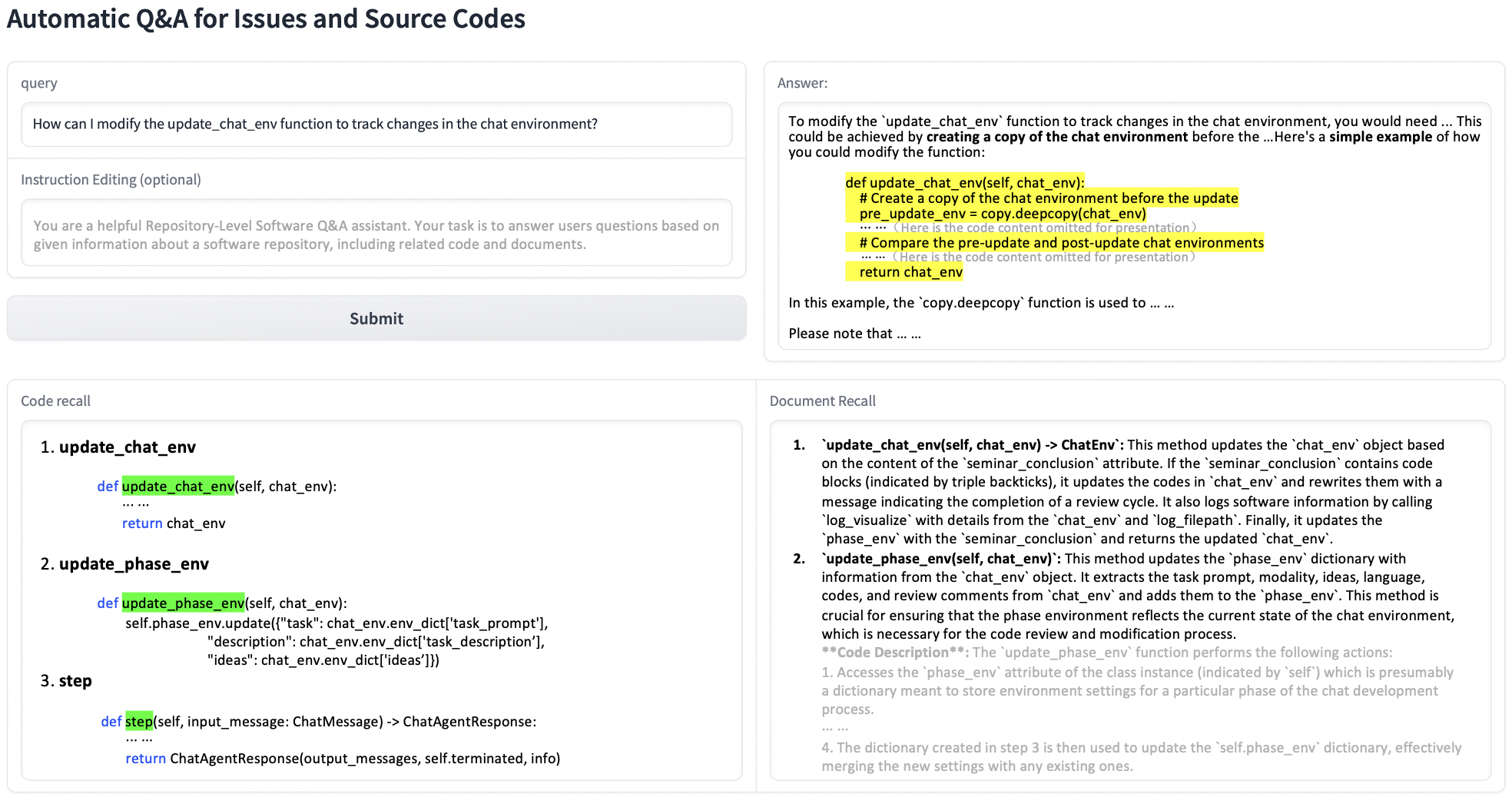} 
\caption{Demonstration of Automatic Q\&A for Issues and Code Explanation by the ``Chat~With~Repo'' interface.} 
  \label{fig:chat_with_repo_ui}
\end{figure*}

Here we demonstrate a preliminary prototype of \textbf{Automatic Q\&A for Issues and Code Explanation}. A running example is shown in \Cref{fig:chat_with_repo_ui}. The program begins by pre-vectorizing code documentation and storing it in a vector database. When a query request is received, it is transformed into an embedding vector for fetching relevant documentation information from the database. This is followed by using the documentation's MetaInfo to locate the pertinent source code, effectively retrieving relevant sections of both documentation text and source code. Moreover, beyond embedding search, a multi-way recall mechanism has been developed, incorporating entity recognition with keyword search. This involves extracting code entities from the user's question using a LLM, and conducting searches across documentation and code repositories to match the top K returned documentation and code blocks. A weighting module has been developed for recalling the most relevant information. Additionally, we input directory tree information to help the model better understand the entire repository. The final step is to concatenate documentation and code blocks retrieved through both mechanisms, along with the target object's parent code, referencing code, and directory tree information, into a prompt for the LLM to generate answers. This sophisticated RAG-based retrieval system bridges human natural language with code language, enabling precise recall at the repository level and paving the way for downstream applications.

\newpage

\noindent A real world ``Chat With Repo'' example with input and output is shown as follows.

\subsection*{Query}
\begin{lstlisting}[label={lst:chat_with_repo_query}]

How would you modify the update\_chat\_env function to track changes in the chat environment?
\end{lstlisting}

\subsection*{Answer}
\begin{lstlisting}[label={lst:chat_with_repo_answer}]
To modify the `update_chat_env` function to track changes in the chat environment, you would need to add a mechanism to store the state of the chat environment before and after the update. This could be achieved by creating a copy of the chat environment before the update and comparing it with the state of the chat environment after the update. Here's a simple example of how you could modify the function:

```python
def update_chat_env(self, chat_env):
    # Create a copy of the chat environment before the update
    pre_update_env = copy.deepcopy(chat_env)

    # Update the chat environment as per your requirements
    # This is where you would add your update logic

    # Compare the pre-update and post-update chat environments
    if pre_update_env != chat_env:
        print("Chat environment has been updated.")
    else:
        print("Chat environment has not been changed.")

    return chat_env
```

In this example, the `copy.deepcopy` function is used to create a deep copy of the `chat_env` object. This ensures that changes to the `chat_env` object after the copy do not affect the `pre_update_env` object. After the update, the `pre_update_env` and `chat_env` objects are compared. If they are not equal, this means that the chat environment has been updated.

Please note that this is a simple example and the actual implementation may vary depending on the specific requirements of your project. For instance, you might want to log the changes in a file instead of printing them, or you might want to use a more sophisticated method for comparing the chat environments.
\end{lstlisting}

\subsection*{Embedding Recall}
\begin{lstlisting}[label={lst:chat_with_repo_embedding_recall}]
1.
**Function Name**: `update_phase_env`

**Function of update_phase_env**: The function `update_phase_env` is designed to update the environment settings of a particular phase in a chat development context by incorporating various elements from a given chat environment object.

**Parameters**: The parameters of this function include:
- `self`: Represents the instance of the class where the `update_phase_env` function resides. It allows the function to access the attributes and methods of the class instance.
- `chat_env`: An object that contains the environment settings for the chat. It is expected to have an attribute `env_dict` which is a dictionary containing keys such as 'task_prompt', 'modality', 'ideas', and 'language'. Additionally, `chat_env` should have methods `get_codes` and `get_requirements` that return specific information related to codes and requirements.

**Code Description**: The `update_phase_env` function performs the following actions:
1. Accesses the `phase_env` attribute of the class instance (indicated by `self`) which is presumably a dictionary meant to store environment settings for a particular phase of the chat development process.
2. Calls the `update` method on `self.phase_env` to modify its contents.
3. Constructs a new dictionary with the following keys and values:
   - `"task"`: Set to the value associated with the `'task_prompt'` key in the `chat_env.env_dict` dictionary.
   - `"modality"`: Set to the value associated with the `'modality'` key in the `chat_env.env_dict` dictionary.
   - `"ideas"`: Set to the value associated with the `'ideas'` key in the `chat_env.env_dict` dictionary.
   - `"language"`: Set to the value associated with the `'language'` key in the `chat_env.env_dict` dictionary.
   - `"codes"`: Obtained by calling the `get_codes` method on the `chat_env` object.
   - `"requirements"`: Obtained by calling the `get_requirements` method on the `chat_env` object.
4. The dictionary created in step 3 is then used to update the `self.phase_env` dictionary, effectively merging the new settings with any existing ones.

**Note**: When using the `update_phase_env` function, it is important to ensure that the `chat_env` object passed as a parameter is properly initialized and contains all the necessary attributes and methods. The `env_dict` attribute must include the keys 'task_prompt', 'modality', 'ideas', and 'language', and the object must have the `get_codes` and `get_requirements` methods implemented. Failure to meet these conditions may result in a `KeyError` or `AttributeError`. Additionally, this function mutates the state of the `self.phase_env` dictionary, so any changes made will persist in the instance of the class where this function is called.
2.
**CodeReviewModification**: The function of the `CodeReviewModification` class is to manage the phase of code review modifications within a chat development environment. It inherits from a base `Phase` class and is designed to update the environment with information relevant to the code review and modification process.

**Attributes**: The attributes of this Class are inherited from the `Phase` class and any that are added or modified within this class.
- `phase_env`: A dictionary attribute inherited from `Phase` that stores environment variables relevant to the current phase.
- `seminar_conclusion`: Although not explicitly defined in the provided code, it is implied that this attribute is used to store the conclusion of a seminar or review session.

**Code Description**: The `CodeReviewModification` class contains two main methods:

1. `update_phase_env(self, chat_env)`: This method updates the `phase_env` dictionary with information from the `chat_env` object. It extracts the task prompt, modality, ideas, language, codes, and review comments from `chat_env` and adds them to the `phase_env`. This method is crucial for ensuring that the phase environment reflects the current state of the chat environment, which is necessary for the code review and modification process.

2. `update_chat_env(self, chat_env) -> ChatEnv`: This method updates the `chat_env` object based on the content of the `seminar_conclusion` attribute. If the `seminar_conclusion` contains code blocks (indicated by triple backticks), it updates the codes in `chat_env` and rewrites them with a message indicating the completion of a review cycle. It also logs software information by calling `log_visualize` with details from the `chat_env` and `log_filepath`. Finally, it updates the `phase_env` with the `seminar_conclusion` and returns the updated `chat_env`.

**Note**: Points to note about the use of the code:
- The `chat_env` parameter is expected to be an object that contains an `env_dict` with keys such as 'task_prompt', 'modality', 'ideas', 'language', and 'review_comments', as well as methods like `get_codes()` and `update_codes()`.
- The `seminar_conclusion` attribute must be set before calling `update_chat_env` as it uses this attribute to update the `chat_env`.
- The `log_visualize` function and `get_info` function are not defined within the provided code snippet, so they should be implemented elsewhere in the project or imported from a module.
- The `ChatEnv` return type suggests that there is a `ChatEnv` class defined elsewhere in the project, which should be used in conjunction with this class.

**Output Example**: Since the methods do not produce a direct output but rather update the state of objects, there is no typical output example. However, after executing the methods, one could expect the `phase_env` and `chat_env` objects to have updated information reflecting the current state of the code review and modification phase.
3.
**chatting**: The function of `chatting` is to conduct a simulated chat session between two roles within a software development environment, with the goal of reaching a conclusion on a specific phase of the project.

**Parameters**:
- `chat_env`: The global chat environment which contains configurations and context for the chat session.
- `task_prompt`: A string representing the user's query or task that needs to be addressed during the chat.
- `assistant_role_name`: The name of the role assumed by the assistant in the chat.
- `user_role_name`: The name of the role assumed by the user initiating the chat.
- `phase_prompt`: A string containing the prompt for the current phase of the chat.
- `phase_name`: The name of the current phase of the chat.
- `assistant_role_prompt`: The prompt associated with the assistant's role.
- `user_role_prompt`: The prompt associated with the user's role.
- `task_type`: An enumeration value representing the type of task being simulated in the chat.
- `need_reflect`: A boolean indicating whether the chat session requires reflection to generate a conclusion.
- `with_task_specify`: A boolean indicating whether the task needs to be specified within the chat.
- `model_type`: An enumeration value indicating the type of language model to be used for generating responses.
- `placeholders`: A dictionary containing placeholders that can be used to fill in the phase environment for generating the phase prompt.
- `chat_turn_limit`: An integer representing the maximum number of turns the chat session can have.

**Code Description**:
The `chatting` function starts by ensuring that the `placeholders` argument is not `None` and that the `chat_turn_limit` is within an acceptable range (1 to 100). It then checks if the roles specified by `assistant_role_name` and `user_role_name` exist within the `chat_env`.

A `RolePlaying` session is initialized with the provided role names, prompts, task type, and model type. The function then begins the chat session by initializing the first user message using the `init_chat` method of the `RolePlaying` session.

The chat session proceeds in turns, where each turn consists of the user sending a message to the assistant and the assistant responding. The messages and responses are generated by interacting with a language model (LLM). The conversation is logged using a `log_visualize` function, which is not defined within the provided code snippet.

During the chat, the function looks for a special `<INFO>` marker in the conversation, which indicates a significant conclusion has been reached. If such a conclusion is found, or if the chat is terminated, the loop ends.

If the `need_reflect` flag is set, the function may call `self_reflection` to generate a conclusion if one has not been reached during the chat session. The reflection is based on the entire conversation history and the context of the phase.

Finally, the function logs the seminar conclusion, extracts the relevant part after the `<INFO>` marker, and returns it as the result of the chat session.

**Note**:
- The function assumes that the `chat_env` has methods `exist_employee` to check for the existence of roles.
- The `RolePlaying` class is used to simulate the chat session and is expected to have methods like `init_chat` and `step`.
- The `log_visualize` function is used for logging purposes but is not defined within the provided code snippet.
- The function raises a `ValueError` if the specified roles are not found within the `chat_env`.
- The `self_reflection` method is used for generating reflections and is assumed to be a member of the same class.

**Output Example**:
If the chat session concludes with a marked conclusion, the function might return something like:
```
"PowerPoint is the best choice for our presentation needs."
```
If the chat session does not reach a marked conclusion but requires reflection, the `self_reflection` method might return:
```
"Yes"
```
In cases where the chat is terminated without a marked conclusion and no reflection is needed, the last message from the assistant might be returned as is.
4.
**Function Name**: execute

**Purpose**: The function `execute` is designed to handle a phase of a chat development environment by updating the environment, checking for module not found errors, resolving them if present, and conducting a chat session if no such errors are found.

**Parameters**:
- `chat_env`: An instance of `ChatEnv`, which represents the current chat environment.
- `chat_turn_limit`: An integer indicating the maximum number of turns allowed in the chat session.
- `need_reflect`: A boolean indicating whether reflection is needed in the chat session.

**Code Description**:
The `execute` function begins by updating the phase environment with the current `chat_env`. It then checks if there is a "ModuleNotFoundError" in the `test_reports` of the `phase_env`. If such an error is present, it attempts to fix the error by calling `chat_env.fix_module_not_found_error` and logs the error for visualization.

The function then uses a regular expression to find all instances of the error message indicating a missing module and constructs a string `pip_install_content` that contains the commands to install the missing modules using `pip`. This string is also logged for visualization.

If no "ModuleNotFoundError" is found, the function proceeds to conduct a chat session by calling the `chatting` method with various parameters such as `chat_env`, `task_prompt`, `need_reflect`, role names, prompts, `chat_turn_limit`, and placeholders. The result of this chat session is stored in `self.seminar_conclusion`.

After handling the error or conducting the chat session, the function updates the chat environment with the potentially modified `chat_env` and returns it.

**Note**:
- The function assumes that the `chat_env` object has the methods `fix_module_not_found_error` and `update_chat_env` implemented.
- The `chatting` method is also assumed to be implemented and is responsible for conducting the chat session.
- The function uses regular expressions to parse error messages, so it is important that the error messages follow the expected format for the regular expressions to work correctly.
- The function logs actions for visualization, which implies that a logging mechanism should be in place for the output to be meaningful.

**Output Example**:
A possible appearance of the code's return value could be an updated `ChatEnv` object with modifications based on the error handling and chat session conducted within the `execute` function. The object would reflect the new state of the chat environment after the execution of this function.

\end{lstlisting}

\subsection*{Key Words Recall}
\begin{lstlisting}[label={lst:chat_with_repo_key_words_recall}]
Key words:
modify, update_chat_env, track

Code:
['\n```python\n    def update_chat_env(self, chat_env):\n        return chat_env\n\n```']
\end{lstlisting}

\subsection*{Code Recall}
\begin{lstlisting}[label={lst:chat_with_repo_code_recall}]
def update_chat_env(self, chat_env) -> ChatEnv:
    chat_env._update_requirements(self.seminar_conclusion)
    chat_env.rewrite_requirements()
    log_visualize(
        "**[Software Info]**:\n\n {}".format(get_info(chat_env.env_dict['directory'], self.log_filepath)))
    return chat_env

def update_chat_env(self, chat_env) -> ChatEnv:
    chat_env.update_codes(self.seminar_conclusion)
    if len(chat_env.codes.codebooks.keys()) == 0:
        raise ValueError("No Valid Codes.")
    chat_env.rewrite_codes("Code Complete #" + str(self.phase_env["cycle_index"]) + " Finished")
    log_visualize(
        "**[Software Info]**:\n\n {}".format(get_info(chat_env.env_dict['directory'], self.log_filepath)))
    return chat_env
\end{lstlisting}

\end{document}